\documentclass[10pt,twocolumn,letterpaper]{article}

\usepackage{iccv}              

\definecolor{iccvblue}{rgb}{0.21,0.49,0.74}
\usepackage[pagebackref,breaklinks,colorlinks,allcolors=iccvblue]{hyperref}

\usepackage{amsmath}
\usepackage{amssymb}
\usepackage{graphicx}
\usepackage{booktabs}
\usepackage{multirow}
\usepackage{array}
\usepackage{colortbl}
\usepackage{fontawesome}
\usepackage{marvosym}
\definecolor{color1}{RGB}{176, 36, 24}
\definecolor{color2}{RGB}{119,185,0}
\definecolor{color3}{RGB}{0, 0, 200}
\def\paperID{2239} 
\def\confName{ICCV}
\def\confYear{2025}

\title{Language Driven Occupancy Prediction}

\author{Zhu Yu$^{1,\dag}$ \quad Bowen Pang$^{1,2,\dag}$ \quad Lizhe Liu$^{2,}$\textsuperscript{\Letter} \quad Runmin Zhang$^1$ \quad Qiang Li$^2$ \\ Si-Yuan Cao$^{3,1}$ \quad Maochun Luo$^{2}$ \quad Mingxia Chen$^{2}$ \quad Sheng Yang$^{2}$ \quad Hui-Liang Shen$^{1,}$\textsuperscript{\Letter} \\
{\normalsize $^1$Zhejiang University \quad $^2$Unmanned Vehicle Dept., CaiNiao Inc., Alibaba Group} 
\\ {\normalsize $^3$Ningbo Global Innovation Center, Zhejiang University} \\
{\tt\small
    \faGithubAlt~\textbf{Project Page:} \href{https://github.com/pkqbajng/LOcc}{\texttt{https://github.com/pkqbajng/LOcc}}}
}

\begin{document}
	\maketitle
	\begin{abstract}
	We introduce \textbf{LOcc}, an effective and generalizable framework for open-vocabulary occupancy (OVO) prediction. Previous approaches typically supervise the networks through coarse voxel-to-text correspondences via image features as intermediates or noisy and sparse correspondences from voxel-based model-view projections. To alleviate the inaccurate supervision, we propose a \textbf{semantic transitive labeling} pipeline to generate dense and fine-grained 3D language occupancy ground truth. Our pipeline presents a feasible way to dig into the valuable semantic information of images, transferring text labels from images to LiDAR point clouds and ultimately to voxels, to establish precise voxel-to-text correspondences. By replacing the original prediction head of supervised occupancy models with a geometry head for binary occupancy states and a language head for language features, LOcc effectively uses the generated language ground truth to guide the learning of 3D language volume. Through extensive experiments, we demonstrate that our transitive semantic labeling pipeline can produce more accurate pseudo-labeled ground truth, diminishing labor-intensive human annotations. Additionally, we validate LOcc across various architectures, where all models consistently outperform state-of-the-art zero-shot occupancy prediction approaches on the Occ3D-nuScenes dataset.
\end{abstract}
    {\renewcommand\thefootnote{}\footnotetext{$\dag$: Equal contribution. \Letter: Corresponding author.}}
	\section{Introduction}
\label{sec:intro}
Vision-based occupancy prediction aims to estimate both the complete scene geometry and semantics using only image inputs, which serves as a critical foundation for various 3D perception tasks, such as autonomous driving~\cite{UniAD}, embodied agent~\cite{embodiedscan, TOD}, mapping and planning~\cite{OccupancySurvey, OccNet}. Existing supervised occupancy prediction approaches~\cite{occ3d, SurroundOcc, OccNet, OpenOccupancy, RenderOcc, COTR} are generally constrained to predicting a fixed set of semantic categories, as the dense occupancy ground truth is generated by merging multi-frame LiDAR point clouds with semantic annotations. In this process, human annotators must label each point across thousands of LiDAR frames with a semantic class, which remains highly labor-intensive and costly. Instead, open-vocabulary occupancy (OVO) facilitates predicting the occupancy with arbitrary sets of vocabularies using only unlabeled image-LiDAR data for training~\cite{pop3d, VEON}.

\begin{figure}
	\centering\footnotesize
	\newcolumntype{P}[1]{>{\centering\arraybackslash}m{#1}}
	\renewcommand{\arraystretch}{0.8}
	\begin{tabular}{P{0.44\linewidth} P{0.44\linewidth}}
		\includegraphics[width=\linewidth]{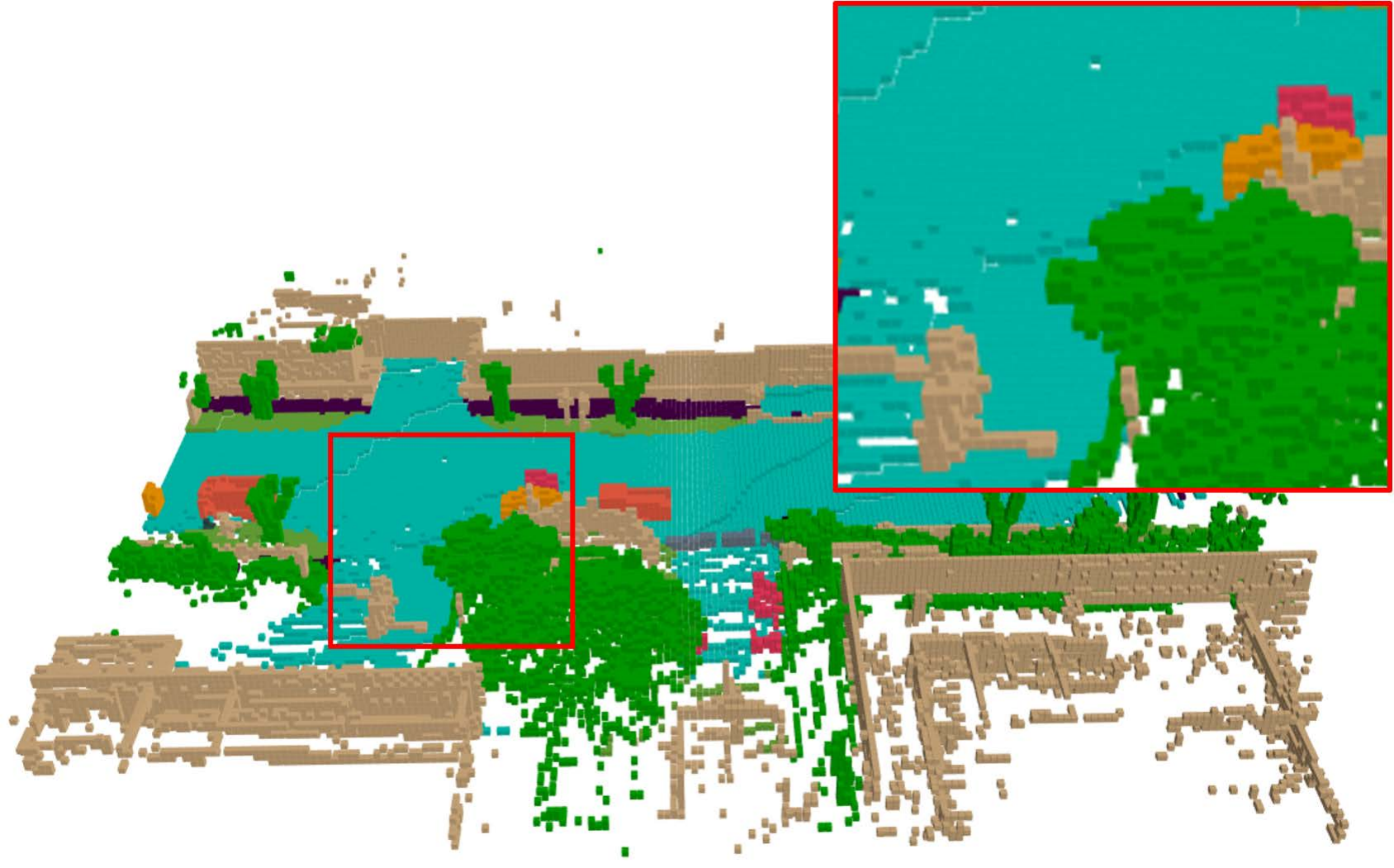} &
		\includegraphics[width=\linewidth]{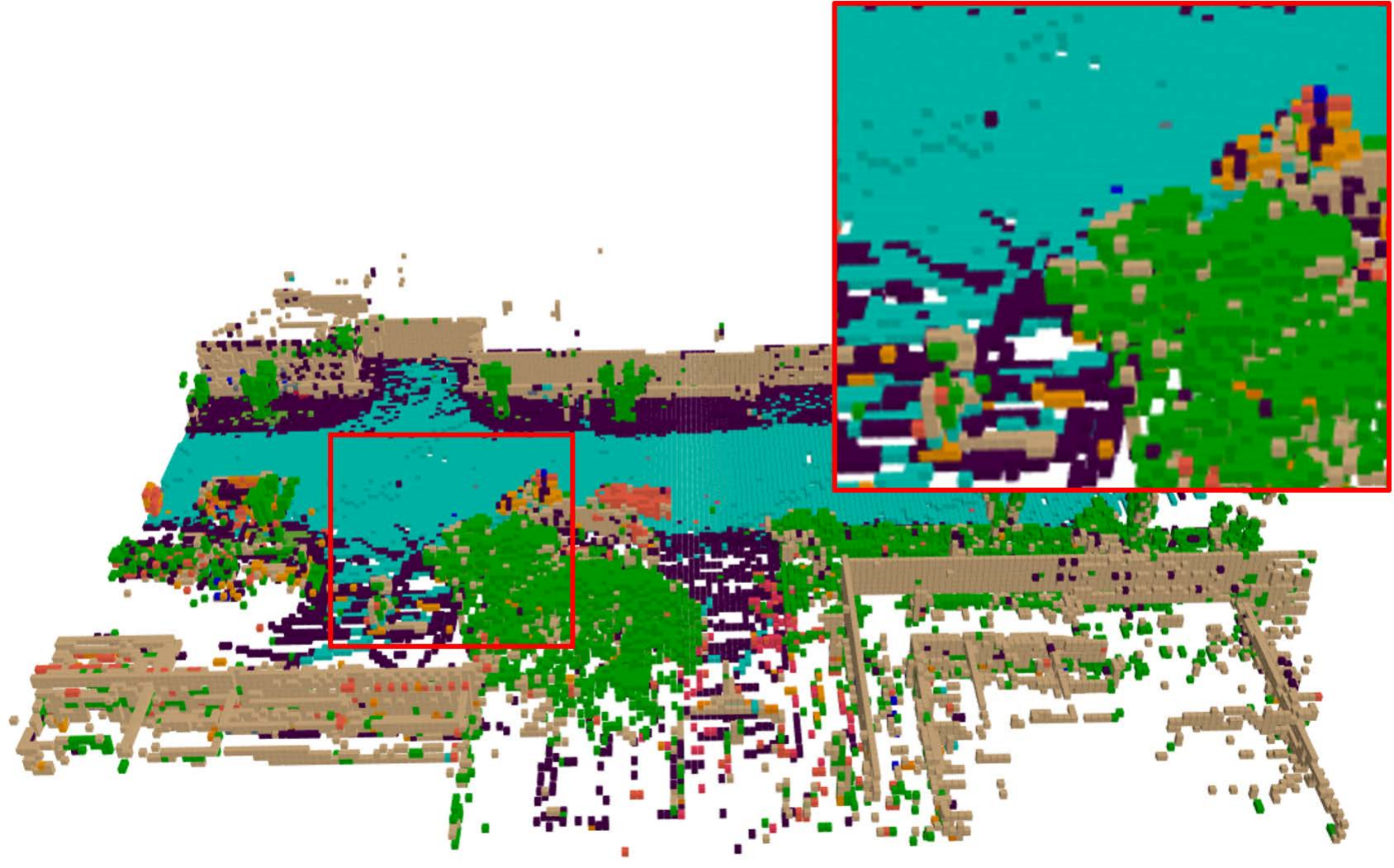} \\
		(a) Human-annotated ground truth & (b) Image features as intermediates\\
		\includegraphics[width=\linewidth]{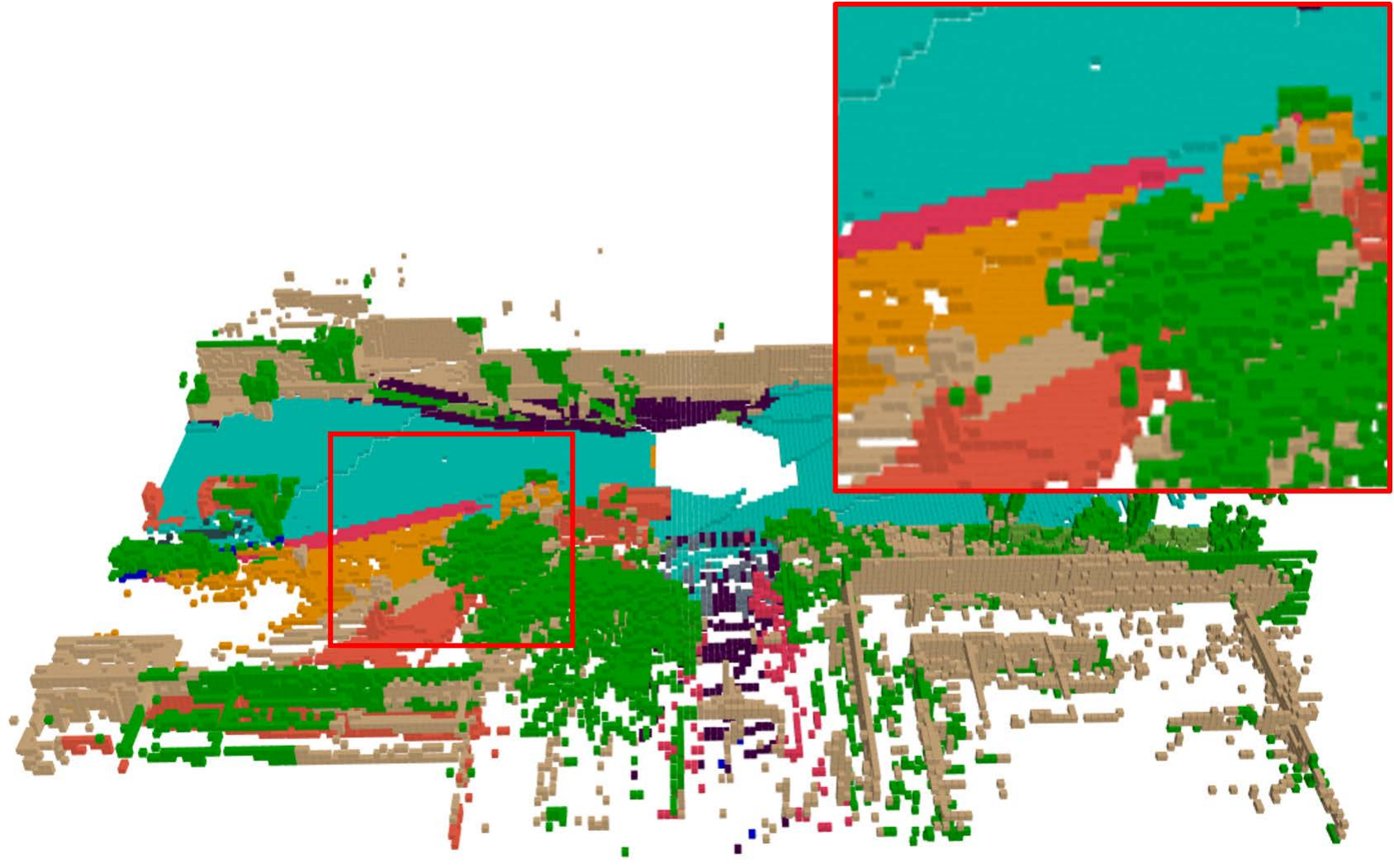} &		
		\includegraphics[width=\linewidth]{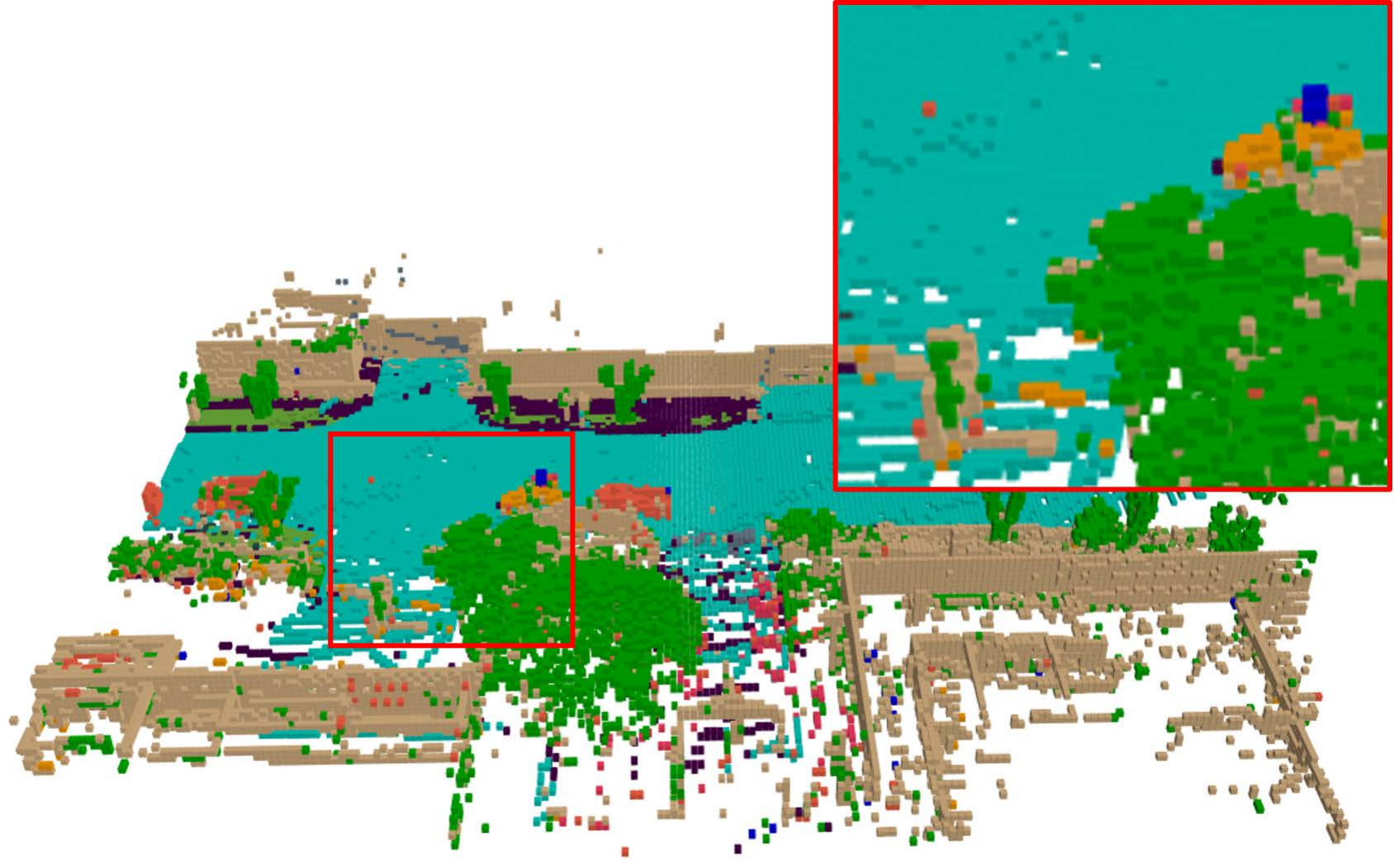} \\
		(c) Voxel-based model-view projection  & (d) Semantic transitive labeling (ours) 
	\end{tabular}
	\caption{Comparison of pseudo-Labeled 3D language occupancy ground truth from different pipelines.} 
	\label{fig:gt_comparison}
	\vspace{-5mm}
\end{figure}

Due to the lack of large-scale occupancy datasets with language annotations, distilling knowledge from vision-language models~\cite{pop3d, CLIP, LSeg, LangOcc} into occupancy prediction networks presents a viable solution. Some approaches~\cite{pop3d, LSeg} use image features from the pretrained vision-language models~\cite{CLIP, LSeg} as intermediates~\cite{LangOcc, pop3d}, subsequently bridging the gap between 3D voxel and language features. However, the feature values of objects with the same semantics may vary, as image features inherently encode both semantic and appearance information, leading to inconsistencies in semantic representations across different inputs. Consequently, the pseudo-labeled ground truth derived from these features is inherently noisy in terms of semantics, failing to provide precise guidance for voxel-level understanding. Moreover, the single-scan sparse LiDAR~\cite{pop3d} can only provide sparse geometry supervision, further resulting in sparse and coarse voxel-to-text correspondences. To address this, VEON~\cite{VEON} enhances geometry supervision by employing dense binary occupancy maps, which are generated through offline post-processing. However, it treats voxels as point clouds and assigns semantic labels to them by model-view projection, which overlooks the occlusion problem and only uses single-frame images, leading to coarse voxel-to-text correspondences, too.

To address the aforementioned issues, we revisit the core aspect of OVO: generating dense and fine-grained 3D language occupancy ground truth. In this work, we propose a \textbf{semantic transitive labeling} pipeline to achieve this goal, without laborious procedures. It can be roughly divided into two steps: label transferring and scene reconstruction. For the first step, we use large vision-language models (LVLM)~\cite{QwenVL, qwen, llava} and open-vocabulary segmentation models (OV-Seg)~\cite{SAN, ODISE, Catseg} to assign text labels to each pixel, obtaining consistent semantic representations.  Then, through the projection of LiDAR points, the text labels are transferred to the LiDAR points, bridging the gap between 3D data and texts. In the reconstruction process, we model the occlusion relationships of the point clouds during point projection, preventing the wrong label assignment of voxel-based model-view projection. With the pseudo-labeled LiDAR data, we reconstruct a temporally dense scene by merging multi-frame LiDAR points to obtain a dense 3D spatial representation.
Furthermore, to reduce the impact of segmentation noise from individual frames, we apply majority-voting voxelization to assign the most frequent label to each voxel. The resulting ground truth enhances 3D language supervision for open-vocabulary occupancy networks. A comparison of different pseudo labeled ground truth from different pipelines is illustrated in Fig.~\ref{fig:gt_comparison}.

Based on the semantic transitive labeling pipeline, we devise \textbf{LOcc}, an effective and generalizable framework that is compatible with most existing supervised occupancy models~\cite{BEVDet, BEVDet4D, bevformer} for OVO. To efficiently align high-dimensional CLIP~\cite{CLIP} embeddings, we devise a language autoencoder that maps them to a low-dimensional latent space, reducing computational cost. We validate our framework on several mainstream approaches, referred to as LOcc-BEVFormer, LOcc-BEVDet, and LOcc-BEVDet4D. All three models consistently surpass all the previous state-of-the-art zero-shot occupancy prediction approaches on the Occ3D-nuScenes dataset. Notably, even the simpler LOcc-BEVDet, with an input resolution of 256$\times$704, achieves a mIoU of 20.29, outperforming previous methods that rely on temporal image inputs, higher-resolution inputs, or larger backbone networks. In summary, our contributions are summarized as follows:
\begin{itemize}
	
	\item We propose LOcc, an effective and generalizable framework that is compatible with most existing supervised methods for OVO. To reduce the computational cost for the language feature alignment, a text-based autoencoder is introduced to map the high-dimensional CLIP embeddings to a low-dimensional latent space.
	
	\item We propose a semantic transitive labeling pipeline for generating dense and fine-grained 3D language occupancy ground truth. This pipeline ensures that text semantic labels can be effectively transferred from images to LiDAR point clouds and ultimately to voxels. We experimentally verify that our pipeline can produce more accurate pseudo-labeled ground truth, diminishing time-consuming human annotations. 
	
	\item We validate our LOcc framework with several mainstream models, which all demonstrate superior performance over previous state-of-the-art methods, proving its generablizability and effectiveness.
\end{itemize}
	\section{Related Work}
\label{sec:relate}

\subsection{Vision-based Occupancy Prediction}
Occupancy prediction jointly estimates the occupancy state and semantic label for every voxel in the scene using images. Voxels are classified as free, occupied, or unobserved, with occupied voxels assigned corresponding semantic labels. Early works~\cite{SemanticKITTI, MonoScene, OccFormer, VoxFormer, MonoOcc, CGFormer} typically use monocular or stereo images as inputs, also referred to as semantic scene completion (SSC). With advancements in 3D perception~\cite{bevformer, LiftSplat, bevformerv2, BEVDet, BEVDet4D, TPVFormer, DFA3D, TPVD, voxdet, pointdc, FlashOcc}, the occupancy of surrounding scenes using multi-camera images~\cite{DHD, seedark, COTR, TPVFormer, CVTOcc} has garnered significant attention due to its strong representation capabilities in 3D space. SurroundOcc~\cite{SurroundOcc} introduces a new pipeline for constructing dense occupancy ground truth. Occ3D~\cite{occ3d} establishes benchmarks for surround occupancy prediction through a three-stage data generation pipeline: voxel densification, occlusion reasoning, and image-guided voxel refinement. OpenOccupancy~\cite{OpenOccupancy} provides a benchmark with higher-resolution ground truth, while OccNet~\cite{OccNet} offers flow annotations. 

\subsection{Open-Vocabulary Understanding}
The recent advances in vision-language models~\cite{CLIP} have shown a remarkable zero-shot performance by using only paired images and captions for model training, demonstrating the strong connection between images and natural language. Then, numerous works explore the potential of language driven downstream applications, such as object detection~\cite{OVOD, YoloWorld}, tracking~\cite{OVTrack}, and segmentation~\cite{LSeg, Catseg, SAN, ODISE}. Taking the image features of these 2D foundation models as intermediaries, OpenScene~\cite{OpenScene} distills 2D vision-language knowledge into a 3D point cloud network and achieves zero-shot 3D segmentation. LERF~\cite{LERF} is the first to embed CLIP features into NeRF for controllable rendering. LangSplat~\cite{LangSplat} uses SAM~\cite{SAM} to tackle the point ambiguity issue for 3D language field modeling.

\begin{figure*}[t]
	\centering\includegraphics[width=\linewidth]{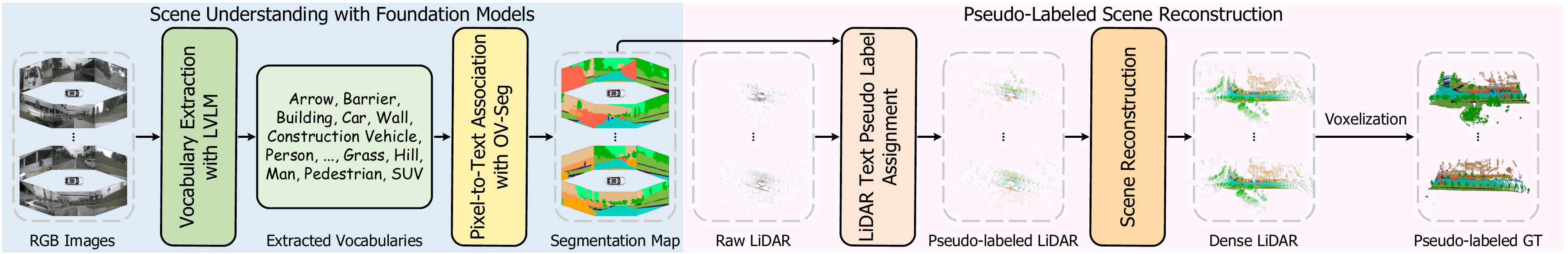}
	\caption{Framework of our semantic transitive labeling pipeline for generating dense and fine-grained pseudo-labeled 3D language occupancy ground truth. Given surround images of the entire scene, we first map them to a vocabulary set using a large vision-language model (LVLM) and associate pixels with these terms through an open-vocabulary segmentation model (OV-Seg). Based on the segmentation results, we assign pseudo labels to the points by projecting them onto the corresponding image plane and merge all the point clouds to form a complete scene. Finally, we voxelize the scene reconstruction results to establish dense and fine-grained voxel-to-text correspondences, generating the pseudo-labeled ground truth.}
	\label{fig:reconstruction}
	\vspace{-5mm}
\end{figure*}

\subsection{Open-Vocabulary Occupancy Prediction}
Benefiting from the development of vision-language foundation models, OVO enables the prediction of occupancy associated with arbitrary vocabulary sets without requiring manual annotations of extensive 3D voxel data. POP-3D~\cite{pop3d} employs only unlabeled images and sparse LiDAR point clouds for training. It distills vision-language knowledge from the pretrained LSeg~\cite{LSeg} into the occupancy networks by computing cosine similarity between 3D features and their corresponding 2D features, obtained by projecting the point cloud onto the image plane. VEON~\cite{VEON} introduces a side adaptor network to integrate CLIP~\cite{CLIP} into the occupancy prediction model. However, the extracted image features may exhibit inconsistencies across different input images, lacking accurate guidance for voxel-level understanding and leading to coarse voxel-to-text correspondences. Furthermore, current works typically treat the voxels as point clouds and employ voxel-based model-view projection to obtain supervision, leading to a lots of misprojections.
	\section{Method}
\label{sec:approach}

\begin{figure}[t]
	\centering\includegraphics[width=\linewidth]{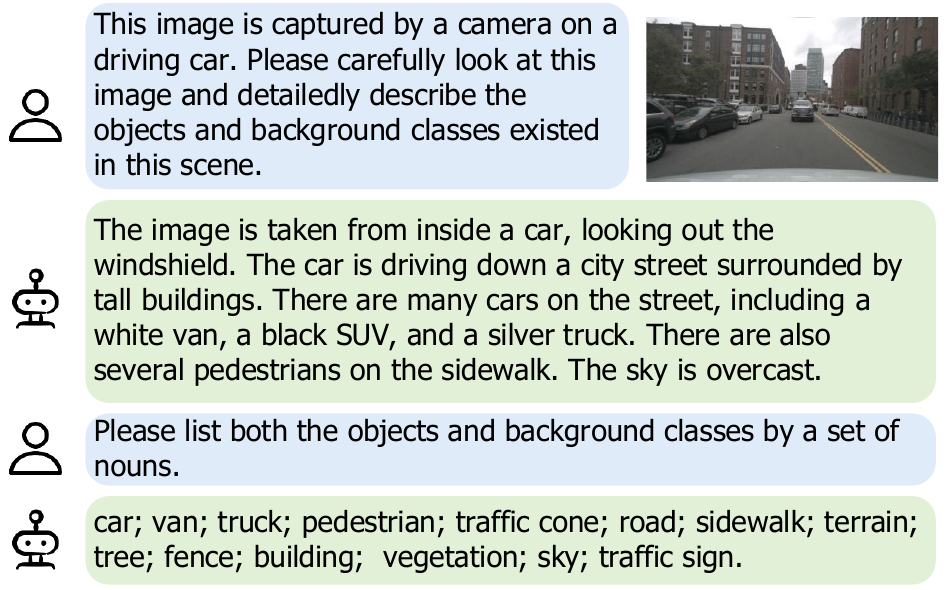}\vspace{-2mm}
	\caption{An example of the conversation process with Qwen-VL~\cite{QwenVL} to extract vocabularies from an individual image.}
	\label{fig:conversion}
	\vspace{-3mm}
\end{figure}

Our proposed LOcc framework mainly consists of two aspects: generating dense and fine-grained pseudo-labeled 3D language occupancy ground truth, diminishing laborious human annotations (Sec.~\ref{sec:gt_generation}), and using the generated ground truth to train the OVO networks (Sec.~\ref{sec:ovo}). To reduce computational cost, an autoencoder is introduced to map the CLIP~\cite{CLIP} embeddings into a lower dimensional latent space (Sec.~\ref{sec:autoencoder}).

\subsection{Pseudo-Labeled Ground Truth Generation}
\label{sec:gt_generation}
We propose a semantic transitive labeling pipeline to generate dense and fine-grained 3D language ground truth. The overview framework is illustrated in Fig.~\ref{fig:reconstruction}. Specifically, we first use vision-language foundation models, including Large Vision-Language Model (LVLM) and Open-Vocabulary Segmentation Model (OV-Seg), to associate each pixel with a text-based pseudo label. We then project the unlabeled LiDAR point clouds onto the image plane to propagate these pseudo labels to each point. The resultant pseudo-labeled LiDAR point clouds can further be used for scene reconstruction. Each voxel subsequently derives its pseudo label from the pseudo-labeled LiDAR points it contains, ultimately generating dense and fine-grained pseudo-labeled language occupancy ground truth.

\textbf{Vocabulary Extraction with LVLM.} Benefiting from the advent of LVLMs~\cite{llava, qwen}, we can decode visual information in images using textual descriptions. Here, we employ the LVLM as an initial perception model to list the objects within each image. 
Fig.~\ref{fig:conversion} illustrates the process of requiring the LVLM to list all classes within an image using Qwen-VL~\cite{QwenVL}. The conversation process follows the chain-of-thought~\cite{ChainofThought} approach. Specifically, rather than directly requesting the LVLM to generate class nouns, we first ask it to provide a description of the scene, followed by a dialogue in which we prompt the LVLM to list the names of all identified classes. To enhance the LVLM's understanding of outdoor environments, we provide an example in advance to assist this process when analyzing a large number of cases. The results extracted from single-frame surround images are then consolidated to represent the overall classes for that frame. 

\begin{figure}[t]
	\centering
	\includegraphics[width=\linewidth]{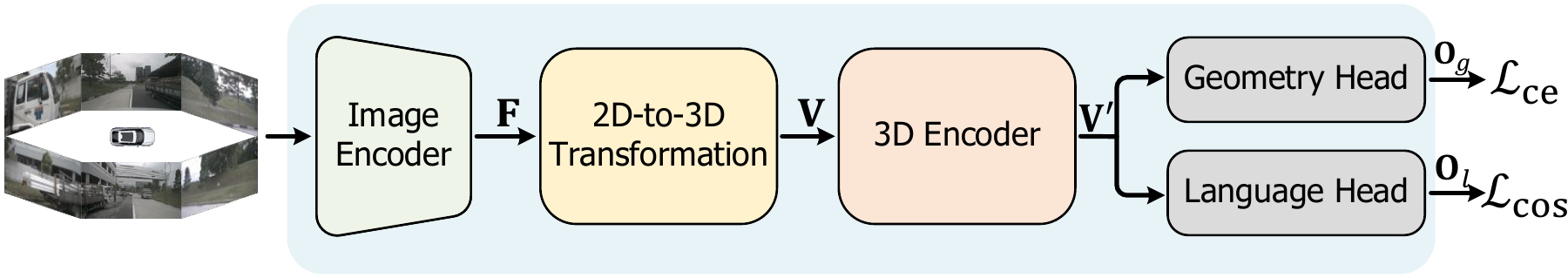}
	\caption{Overview of the open-vocabulary occupancy (OVO) model architecture. }
	\label{fig:OVO_architecture}
	\vspace{-5mm}
\end{figure}

\textbf{Pixel-to-Text Association with OV-Seg.} Having obtained the vocabularies within a frame of surround images, we can establish pixel-to-text association with open-vocabulary semantic segmentation models~\cite{Catseg, ODISE, SAN}. This process begins by mapping the image and vocabularies into a shared feature space via the feature-aligned vision encoder $\Phi_{I}$ and text encoder $\Phi_{T}$, respectively. Each pixel is then assigned a text label by computing cosine similarity with the set of vocabulary embeddings, with the highest-scoring text designated as the label for that pixel.  Specifically, for the $k$-th frame surround-view images $\mathcal{I}_{k} = \left\{ \mathbf{I}_{i, k} \in \mathbb{R}^{H \times W \times 3}, i \in \{ 1, \ldots, N_{c} \} \right\}$ and vocabularies $\mathcal{T}_{k}=\left\{\mathbf{T}_{j,k}, j\in\left\{1, ..., N_{t, k}\right\}\right\}$, their corresponding segmentation maps $\mathcal{S}_{k}=\left\{\mathbf{S}_{i,k}\in\mathbb{R}^{H\times W}, i\in\left\{1, ..., N_{c}\right\}\right\}$ can be obtained as
\begin{gather}
	\mathbf{F}_{i,k} = \Phi_{I}\left(\mathbf{I}_{i,k}\right)\in\mathbb{R}^{H\times W\times C}, \mathbf{E}_{j,k} = \Phi_{T}\left(\mathbf{T}_{j,k}\right)\in\mathbb{R}^{C} \notag, \\
	\mathbf{S}_{i, k}(x, y) = \mathop{\arg\max}_{\mathbf{T}_{j,k}} \text{cos}(\mathbf{F}_{i, k}(x, y), \mathbf{E}_{j, k}),
	\label{eq:open_seg}
\end{gather}
where $\mathbf{F}_{i}$ represents the feature maps of images and $\mathbf{E}_{j}$ represents the embeddings of texts. Here, $N_{t, k}$ indicates the length of the vocabulary set for the $k$-th frame. Typically, $\Phi_{I}$ generates feature maps at a lower resolution and the final high-resolution segmentation maps are produced using an upsampling decoder~\cite{Catseg}. For simplicity, we let $\mathbf{F}_{i}$ have the same resolution with the original image.

In our investigation, we found that the LVLM occasionally fails, overlooking many important classes and listing only a limited vocabulary set within a single frame. To address this issue, we treat multiple frames as a unified sequence, integrating the vocabularies from each frame into a cohesive set $ \mathcal{T}=\left\{{\mathbf{T}_j}, j \in \left\{1,..., N_{t}\right\}\right\}$, where $N_{t}=\sum_{k}^{K} N_{t,k}$. This integrated set is then used to perform open-vocabulary segmentation as described in Eq.~\ref{eq:open_seg}. The key motivation behind this approach is the overlap between consecutive frames, where recurring objects display a degree of similarity. Consequently, frames with incomplete text classes can be supplemented by the outputs from adjacent frames within the same sequence. 

\textbf{LiDAR Text Pseudo Label Assignment.} Using the generated segmentation maps, we can assign text pseudo labels to the unlabeled LiDAR $\mathcal{P}=\left\{\mathbf{P}_{k}, k\in\left\{1, ..., K\right\}\right\}$. Specifically, for a point $\mathbf{P}_{k}(p)$ in the $k$-th frame point cloud, its corresponding image coordinates on the $N_{c}$ surround images are defined as
\begin{equation}
	\left(x_{i}, y_{i}, z_{i}\right) = \mathbf{K}_{i}\left(\mathbf{R}_{i}\mathbf{P}_{k}(p) + \mathbf{t}_{i}\right), i\in\left\{1, ..., N_{c}\right\},
\end{equation}
where $\mathbf{K}_{i}, \mathbf{R}_{i}, \mathbf{t}_{i}$ are camera intrinsic and extrinsic parameters. Among these coordinates, the one that lies within the image boundaries and has the smallest depth value $z_{i}$ serves as the pixel corresponding to point $\mathbf{P}_{k}(p)$. This pixel is then used to sample the segmentation map to obtain the pseudo label $\hat{\mathbf{S}}(p)$ for $\mathbf{P}_{k}(p)$. This process can be summarized as
\begin{gather}
	n = \mathop{\arg\min}_{m} z_{m}, 0 < x_m < W, 0< y_m < H, \notag \\
	\hat{\mathbf{S}}(p) = \phi\left(\mathbf{S}_{n, k}, \left(x_n, y_n\right)\right),
\end{gather}
where $\phi$ indicates the nearest neighbor sampling of $\mathbf{S}_{n, k}$ at the coordinate $\left(x_n, y_n\right)$.

\begin{figure}[t]
	
	\centering
	\includegraphics[width=\linewidth]{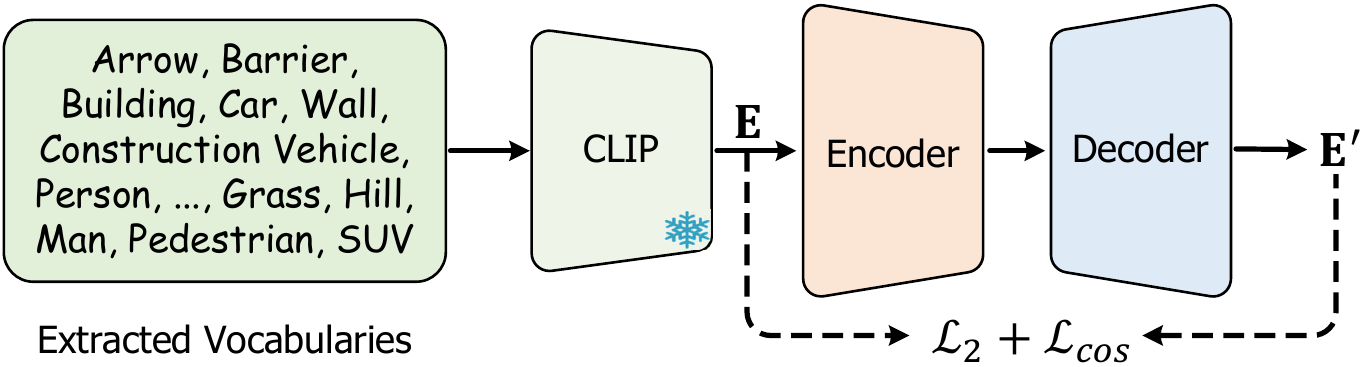}
	\caption{Framework of the autoencoder to map the language features into a low-dimensional latent space.}
	\label{fig:autoencoder}
	\vspace{-5mm}
\end{figure}

\textbf{Scene Reconstruction.} The pseudo-labeled LiDAR data is then used to perform scene reconstruction to establish correspondences between the texts and the voxels. To combine the multi-frame point clouds, we first transform their coordinates into the world coordinate system using the calibration matrices and ego poses. In alignment with VEON~\cite{VEON}, we distinguish moving and static objects by leveraging the geometric information of 3D bounding boxes. For the $k$-th frame, we transform the unified point cloud to this frame and apply majority-voting voxelization on it to obtain the pseudo-labeled 3D language occupancy ground truth. Each voxel takes the most frequent text within it as its text label, and voxels without point clouds are set to empty. Compare to the voxel-based model-view projection, our method obtains text label by projecting LiDAR point clouds onto image planes, which are easily calibrated with the corresponding images during the data collection process. Besides, the majority-voting is more robust and less susceptible to the influence of single-point projection errors, generating ground truth with higher metrics compared to the nearest-point voxelization, as shown in the ablation experiments.

\subsection{Open Vocabulary Occupancy Prediction}
\label{sec:ovo}
Based on the processes described above, we successfully establish dense and fine-grained voxel-to-text correspondences. By minimally modifying most existing supervised occupancy methods~\cite{BEVDet, BEVDet, bevformer}, we can leverage the obtained pseudo-labeled 3D language occupancy ground truth to train a robust OVO model, without the need for additional complex network designs. The overall network architecture is shown in Fig.~\ref{fig:OVO_architecture}. Similar to most existing occupancy network structures, it involves an image encoder, a 2D-to-3D view transformation module, a 3D encoder, except the original prediction head is replaced by a geometry head to predict the binary occupancy state and a language head to estimate the language features.

\textbf{Image Encoder.} The image encoder includes an image backbone network for extracting multi-scale features, complemented by an FPN~\cite{secondfpn} network to enhance features across different levels. The enhanced features $\mathbf{F}\in{\mathbb{R}^{H\times W\times C}}$ are then used to construct 3D features.

\textbf{2D-to-3D Transformation.} The 2D-to-3D view transformation aims to project multiview image features into the 3D voxel space. This can be achieved by either forward projection~\cite{LiftSplat} or backward projection~\cite{bevformer}. In this work, we verify the effectiveness of our proposed method on both projection approaches~\cite{BEVDet4D, BEVDet, bevformer}. The constructed 3D volume $\mathbf{V}$ has a shape of $X\times Y\times Z$ for voxel representation or $X\times Y$ when pooled into BEV representation

\textbf{3D Encoder.} After constructing the 3D volume $\mathbf{V}$, the 3D encoder further refines its representation by performing 3D convolution~\cite{ResNet} or stacking self-attention layers~\cite{bevformer} to obtain the fine-grained volume $\mathbf{V}'$. This refined volume $\mathbf{V}'$ is then fed to a geometry head to estimate binary occupancy states $\mathbf{O}_{g}$ and a language head to generate 3D language volume $\mathbf{O}_{l}$.

\textbf{Training and Evaluation.} During training, we jointly supervise the geometry and language heads to optimize the neural network. Given the binary occupancy ground truth $\hat{\mathbf{O}}_{g}$, we adopt cross entropy loss to supervise $\mathbf{O}_{g}$,
\begin{equation}
	\mathcal{L}_{ce} = \text{CrossEntropy} (\hat{\mathbf{O}}_{g}, \mathbf{O}_{g}).
\end{equation}
Given the pseudo-labeled 3D language ground truth, each voxel of it corresponds to a text. We perform alignment between $\mathbf{O}_{l}$ and the CLIP embeddings of these texts $\hat{\mathbf{O}}_{l}$ using cosine similarity,
\begin{equation}
	\mathcal{L}_{cos}=\sum_{p\in{\mathcal{C}(\hat{\mathbf{O}}_{g})}}1-\text{cos}(\hat{\mathbf{O}}_{l}(p), \mathbf{O}_{l}(p)),
\end{equation}
where $\mathcal{C}(\hat{\mathbf{O}}_{g})$ denotes the set of occupied voxels in $\hat{\mathbf{O}}_{g}$. The final loss function $\mathcal{L}$ is a combination of $\mathcal{L}_{\text{ce}}$ and $\mathcal{L}_{\text{cos}}$,
\begin{equation}
	\mathcal{L} = \mathcal{L}_{\text{ce}} + \mathcal{L}_{\text{cos}}.
\end{equation}

During evaluation, $\mathbf{O}_{g}$ first defines the set of occupied voxels. Then for each occupied voxel, its corresponding language feature is used to compute cosine similarity with the predefined text features, where the text with the highest score represents its semantics.

\subsection{Language Autoencoder}
\label{sec:autoencoder}
Generally, the dimension of CLIP features is 512. To reduce the memory requirements, we introduce a language autoencoder to map the language features into a low-dimensional latent space. The extracted vocabularies of the entire dataset contains approximately one thousand unique classes, allowing the autoencoder to be trained efficiently. Fig.~\ref{fig:autoencoder} shows the framework of the autoencoder. 

Specifically, given the vocabulary set $\mathcal{T}_{a}=\left\{\mathbf{T}_{i},i\in{1,...,N_{a}} \right\}$ of the entire dataset, the encoder $\Phi_{E}$ maps the $D$-dimensional CLIP features $\mathbf{E}\in\mathbb{R}^{D}$ to $\mathbf{E}'\in\mathbb{R}^{D'}$, where $D'<D$. Then we learn a decoder $\Phi_{D}$ to reconstruct the original embedding from the compressed $\mathbf{E}^{'}$. The autoencoder is trained with a combination of L2 loss and cosine loss,
\begin{equation}
	\mathcal{L}_{a} = ||\mathbf{E} - \hat{\mathbf{E}}||_{2} + \text{cos}(\mathbf{E}, \hat{\mathbf{E}}), 
\end{equation}
where $\hat{\mathbf{E}}$ denotes the reconstructed embedding. After training the autoencoder, the language volume of the OVO model is aligned with the compressed embedding.
	\definecolor{nbarrier}{RGB}{112, 128, 144}
\definecolor{nbicycle}{RGB}{220, 20, 60}
\definecolor{nbus}{RGB}{255, 127, 80}
\definecolor{ncar}{RGB}{255, 158, 0}
\definecolor{nconstruct}{RGB}{233, 150, 70}
\definecolor{nmotor}{RGB}{255, 61, 99}
\definecolor{npedestrian}{RGB}{0, 0, 230}
\definecolor{ntraffic}{RGB}{47, 79, 79}
\definecolor{ntrailer}{RGB}{255, 140, 0}
\definecolor{ntruck}{RGB}{255, 99, 71}
\definecolor{ndriveable}{RGB}{0, 207, 191}
\definecolor{nother}{RGB}{175, 0, 75}
\definecolor{nsidewalk}{RGB}{75, 0, 75}
\definecolor{nterrain}{RGB}{112, 180, 60}
\definecolor{nmanmade}{RGB}{222, 184, 135}
\definecolor{nvegetation}{RGB}{0, 175, 0}
\definecolor{nothers}{RGB}{0, 0, 0}

\begin{table*}
	\footnotesize
	\caption{Zero-shot occupancy prediction results on the Occ3D-nuScenes dataset. We report the mean IoU (mIoU) for semantics across different categories, along with per-class semantic IoUs. The symbol $^*$ indicates reproduced results from \cite{CVTOcc}. The methods marked with $\dagger$ indicate that they specifically design the set of vocabularies for training. The methods marked with $\ddagger$ are implemented the official open-source codebase. The top three results among the methods supervised by pseudo-labeled ground truth are in \textcolor{color1}{\textbf{red}}, \textcolor{color2}{green}, and \textcolor{color3}{blue}, respectively.}
	\begin{center}
		\setlength{\tabcolsep}{0.0057\linewidth}
		\resizebox{\textwidth}{!}{\begin{tabular}{l|c|c|c| c c c c c c c c c c c c c c c c c}
				\toprule
				Model
				& Image Backbone & Image Size
				& mIoU
				& \rotatebox{90}{\textcolor{nothers}{$\blacksquare$} others}
				
				& \rotatebox{90}{\textcolor{nbarrier}{$\blacksquare$} barrier}
				
				& \rotatebox{90}{\textcolor{nbicycle}{$\blacksquare$} bicycle}
				
				& \rotatebox{90}{\textcolor{nbus}{$\blacksquare$} bus}
				
				& \rotatebox{90}{\textcolor{ncar}{$\blacksquare$} car}
				
				& \rotatebox{90}{\textcolor{nconstruct}{$\blacksquare$} const. veh.}
				
				& \rotatebox{90}{\textcolor{nmotor}{$\blacksquare$} motorcycle}
				
				& \rotatebox{90}{\textcolor{npedestrian}{$\blacksquare$} pedestrian}
				
				& \rotatebox{90}{\textcolor{ntraffic}{$\blacksquare$} traffic cone}
				
				& \rotatebox{90}{\textcolor{ntrailer}{$\blacksquare$} trailer}
				
				& \rotatebox{90}{\textcolor{ntruck}{$\blacksquare$} truck}
				
				& \rotatebox{90}{\textcolor{ndriveable}{$\blacksquare$} drive. suf.}
				
				& \rotatebox{90}{\textcolor{nother}{$\blacksquare$} other flat}
				
				& \rotatebox{90}{\textcolor{nsidewalk}{$\blacksquare$} sidewalk}
				
				& \rotatebox{90}{\textcolor{nterrain}{$\blacksquare$} terrain}
				
				& \rotatebox{90}{\textcolor{nmanmade}{$\blacksquare$} manmade}
				
				& \rotatebox{90}{\textcolor{nvegetation}{$\blacksquare$} vegetation}
				\\
				\midrule
				\multicolumn{21}{l}{\textit{Supervised by human-labeled ground truth.}}  \\
				\midrule
				MonoScene~\cite{MonoScene} & ResNet-101 & 900 $\times$ 1600 & 6.06 & 1.75 & 7.23 & 4.26 & 4.93 & 9.38 & 5.67 & 3.98 & 3.01 & 5.90 & 4.45 & 7.17 & 14.91 & 6.32 & 7.92 & 7.43 & 1.01 & 7.65 \\
				OccFormer~\cite{OccFormer} & ResNet-101 & 900 $\times$ 1600 & 21.93 & 5.94 & 30.29 & 12.32 & 34.40 & 39.17 & 14.44 & 16.45 & 17.22 & 9.27 & 13.90 & 26.36 & 50.99 & 30.96 & 34.66 & 22.73 & 6.76 & 6.97 \\
				TPVFormer~\cite{TPVFormer} & ResNet-101 & 900 $\times$ 1600  & 27.83 & 7.22 & 38.90 & 13.67 & 40.78 & 45.90 & 17.23 & 19.99 & 18.85 & 14.30 & 26.69 & 34.17 & 55.65 & 35.47 & 37.55 & 30.70 & 19.40 &16.78 \\
				CTF-Occ~\cite{occ3d} & ResNet-101 & 900 $\times$ 1600  & 28.5 & 8.09 & 39.33 & 20.56 & 38.29 & 42.24 & 16.93 & 24.52 & 22.72 & 21.05 & 22.98 & 31.11 & 53.33 & 33.84 & 37.98 & 33.23 & 20.79 & 18.0 \\
				BEVFormer wo TSA$^*$~\cite{bevformer} & ResNet-101 & 900 $\times$ 1600 & 38.05 & 9.11 & 45.68 & 22.61 & 46.19 & 52.97 & 20.27 & 26.5 & 26.8 & 26.21 & 32.29 & 37.58 & 80.5 & 40.6 & 49.93 & 52.48 & 41.59 & 35.51 \\
				BEVFormer$^*$~\cite{bevformer} & ResNet-101 & 900 $\times$ 1600  & 39.04 & 9.57 & 47.13 & 22.52 & 47.61 & 54.14 & 20.39 & 26.44 & 28.12 & 27.46 & 34.53 & 39.69 & 81.44 & 41.14 & 50.79 & 54.00 & 43.08 & 35.60 \\
				CVT-Occ$^*$~\cite{CVTOcc} & ResNet-101 & 900 $\times$ 1600  & 40.34 & 9.45 & 49.46 & 23.57 & 49.18 & 55.63 & 23.10 & 27.85 & 28.88 & 29.07 & 34.97 & 40.98 & 81.44 & 40.92 & 51.37 & 54.25 & 45.94 & 39.71 \\
				BEVDet~\cite{BEVDet} & ResNet-50 & 256 $\times$ 704  & 33.62 & 8.18 & 38.46 & 13.75 & 41.10 & 45.17 & 18.99 & 18.39 & 19.52 & 19.36 & 29.48 & 31.77 & 79.42 & 37.55 & 49.08 & 51.70 & 36.94 & 32.63 \\
				BEVDet4D~\cite{BEVDet4D} & ResNet-50 & 256 $\times$ 704  & 39.11 & 11.03 & 46.68 & 21.59 & 44.13 & 50.70 & 25.30 & 24.12 & 25.89 & 25.21 & 34.13 & 38.94 & 81.33 & 39.08 & 51.96 & 56.13 & 47.41 & 41.15 \\
				\midrule
				\multicolumn{21}{l}{\textit{Supervised by images.}}  \\
				\midrule
				SelfOcc (BEV)~\cite{SelfOcc} & ResNet-50 & 384 $\times$ 800  & 6.76 & 0.00 & 0.00 & 0.00 & 0.00 & 9.82 & 0.00 & 0.00 & 0.00 & 0.00 & 0.00 & 6.97 & 47.03 & 0.00 & 18.75 & 16.58 & 11.93 & 3.81 \\
				SelfOcc (TPV)~\cite{SelfOcc} & ResNet-50 & 384 $\times$ 800  & 9.30 & 0.00 & 0.15 & 0.66 & 5.46 & 12.54 & 0.00 & 0.80 & 2.10 & 0.00 & 0.00 & 8.25 & 55.49 & 0.00 & 26.30 & 26.54 & 14.22 & 5.60 \\
				OccNeRF~\cite{OccNeRF} & ResNet-101 & 928 $\times$ 1600 & 10.81 & 0.00 & 0.83 & 0.82 & 5.13 & 12.49 & 3.50 & 0.23 & 3.10 & 1.84 & 0.52 & 3.90 & 52.62 & 0.00 & 20.81 & 24.75 & 18.45 & 13.19 \\
				\midrule
				\multicolumn{21}{l}{\textit{Supervised by images and unlabeled LiDAR.}}  \\
				\midrule
				VEON-B$\dagger$~\cite{VEON} & ViT-B & 256 $\times$ 704 & 12.38 & \textcolor{color2}{0.50} & 4.80 & 2.70 & 14.70 & 10.90 & \textcolor{color2}{11.00} & 3.80 & 4.70 & 4.00 & \textcolor{color2}{5.30} & 9.60 & 46.50 & \textcolor{color1}{\textbf{0.70}} & 21.10 & 22.10 & 24.80 & 23.70 \\
				VEON-L$\dagger$~\cite{VEON} & ViT-L & 512 $\times$ 1408 & 15.14 & \textcolor{color1}{\textbf{0.90}} & 10.40 & \textcolor{color1}{\textbf{6.20}} & 17.7 & 12.7 & 8.50 & \textcolor{color2}{7.60} & 6.50 & \textcolor{color3}{5.50} & \textcolor{color1}{\textbf{8.20}} & 11.80 & 54.50 & \textcolor{color2}{0.40} & 25.50 & 30.20 & 25.40 & 25.40 \\
				POP-3D-BEVFormer~\cite{pop3d} & ResNet-101 & 256 $\times$ 704 & 11.24 & 0.00 & 7.32 & 0.95 & 11.23 & 9.18 & 2.59 & 2.44 & 2.21 & 0.78 & 1.7 & 8.26 & 56.02 & 0.00 & 19.28 & 21.47 & 21.54 & 26.18 \\
				POP-3D-BEVFormer~\cite{pop3d} & ResNet-101 & 512 $\times$ 1408 & 12.68 & 0.00 & 9.43 & 2.43 & 15.21 & 10.39 & 3.54 & 3.78 & 3.98 & 2.91 & 1.96 & 10.01 & 57.28 & 0.00 & 21.01 & 23.07 & 22.67 & 27.87 \\
				POP-3D-BEVDet~\cite{pop3d} & ResNet-50 & 256 $\times$ 704 & 12.30 & 0.00 & 8.72 & 1.94 & 13.11 & 11.48 & 3.12 & 3.29 & 3.47 & 1.78 & 1.21 & 9.43 & 56.81 & 0.00 & 20.17 & 23.29 & 23.33 & 28.12 \\
				POP-3D-BEVDet~\cite{pop3d} & ResNet-50 & 512 $\times$ 1408 & 12.85 & 0.00 & 7.43 & 2.21 & 15.13 & 12.25 & 3.40 & 3.47 & 4.21 & 2.22 & 2.42 & 10.67 & 58.23 & 0.00 & 21.46 & 24.33 & 22.16 & 28.88 \\
				POP-3D-BEVDet4D~\cite{pop3d} & ResNet-50 & 256 $\times$ 704 & 14.57 & 0.00 & 9.94 & 3.09 & 15.83 & 13.51 & 4.38 & 3.57 & 4.64 & 3.84 & 2.21 & 12.20 & 59.71 & 0.00 & 26.60 & 28.16 & 27.88 & 32.20 \\
				POP-3D-BEVDet4D~\cite{pop3d} & ResNet-50 & 512 $\times$ 1408 & 15.64 & 0.00 & 10.06 & 3.63 & 16.74 & 14.32 & 5.38 & 4.01 & 5.14 & 3.52 & 2.46 & 15.20 & 61.13 & 0.00 & 28.82 & 30.16 & 31.92 & 33.31 \\
				\midrule
				\rowcolor{gray!20}LOcc-BEVFormer (ours) & ResNet-101 & 256 $\times$ 704 & 18.62 & 0.00 & 14.21 & 0.90 & 27.21 & 32.61 & 3.09 & 2.06 & 8.67 & 1.74 & 1.96 & 21.45 & 71.22 & 0.00 & 38.18 & 33.32 & 30.68 & 29.32 \\
				\rowcolor{gray!20}LOcc-BEVFormer (ours) & ResNet-101 & 512 $\times$ 1408 & 19.85 & 0.00 & \textcolor{color3}{14.65} & 1.01 & 27.99 & 35.34 & 5.61 & 4.16 & \textcolor{color3}{11.70} & 3.05 & 2.72 & 23.08 & 72.00 & 0.00 & 39.35 & 34.18 & 32.03 & 30.66 \\
				\rowcolor{gray!20}LOcc-BEVDet (ours) & ResNet-50 & 256 $\times$ 704 & 20.29 & 0.00 & 14.03 & 2.81 & \textcolor{color3}{31.75} & 34.55 & 7.52 & 3.94 & 8.99 & 3.64 & 3.04 & 25.58 & 71.43 & 0.01 & 38.74 & \textcolor{color3}{35.62} & 32.61 & 30.68 \\
				\rowcolor{gray!20}LOcc-BEVDet (ours) & ResNet-50 & 512 $\times$ 1408 & \textcolor{color3}{20.84} & 0.00 & 13.85 & 3.28 & \textcolor{color2}{31.85} & \textcolor{color3}{37.09} & 8.22 & \textcolor{color3}{4.84} & 10.78 & 3.24 & 2.47 & \textcolor{color3}{26.95} & \textcolor{color3}{72.35} & 0.00 & \textcolor{color3}{39.58} & 35.29 & \textcolor{color3}{33.35} & \textcolor{color3}{31.23} \\
				\rowcolor{gray!20}LOcc-BEVDet4D (ours) & ResNet-50 & 256 $\times$ 704 & \textcolor{color2}{22.95} & 0.00 & \textcolor{color2}{15.01} & \textcolor{color3}{5.76} & 29.86 & \textcolor{color2}{38.90} & \textcolor{color1}{\textbf{11.61}} & 4.61 & \textcolor{color2}{13.46} & \textcolor{color2}{6.99} & 2.64 & \textcolor{color1}{\textbf{30.69}} & \textcolor{color2}{73.61} & 0.00 & \textcolor{color2}{40.38} & \textcolor{color1}{\textbf{39.16}} & \textcolor{color2}{38.80} & \textcolor{color2}{38.68} \\
				\rowcolor{gray!20}LOcc-BEVDet4D (ours) & ResNet-50 & 512 $\times$ 1408 & \textcolor{color1}{\textbf{23.84}} & 0.00 & \textcolor{color1}{\textbf{16.92}} & \textcolor{color2}{5.89} & \textcolor{color1}{\textbf{32.94}} & \textcolor{color1}{\textbf{40.08}} & \textcolor{color3}{10.18} & \textcolor{color1}{\textbf{9.85}} & \textcolor{color1}{\textbf{17.11}} & \textcolor{color1}{\textbf{7.21}} & \textcolor{color3}{3.10} & \textcolor{color2}{30.48} & \textcolor{color1}{\textbf{74.13}} & 0.00 & \textcolor{color1}{\textbf{41.25}} & \textcolor{color2}{36.97} & \textcolor{color1}{\textbf{39.57}} & \textcolor{color1}{\textbf{39.66}} \\
				\bottomrule
			\end{tabular}
		}
	\end{center}
	\label{tab:zero_shot}
	\vspace{-5mm}
\end{table*}
\begin{table*}[t]
	\centering
	\Large
	\caption{Ablation study on the framework for generating dense and fine-grained pseudo-labeled 3D language occupancy ground truth. For label transferring process, we analyze three settings: (a) using image features from vision-language foundation models as intermediates, (b) using single-frame vocabularies, and (c) integrating vocabularies across multiple consecutive frames. For scene reconstruction process, we replace the majority-voting voxelization with (d) nearest-point voxelization and (e) voxel-based model-view projection. The best results for each model are in \textbf{bold}.}
	\resizebox{\linewidth}{!}{
		\begin{tabular}{l|ccc|ccc|cccc}
			\toprule
			\multirow{3}{*}{Setting} & \multicolumn{3}{c|}{Label Transferring Process} & \multicolumn{3}{c|}{Scene Reconstruction Process} & \multicolumn{4}{c}{mIoU} \\\cline{2-11}
			& \multirow{2}{*}{Image Feat.} & \multirow{2}{*}{Single-frame vocab.} & \multirow{2}{*}{Consecutive-frame vocab.} &  \multicolumn{2}{c|}{Point-based Projection} & \multirow{2}{*}{Voxel-based Projection} & \multirow{2}{*}{LOcc-BEVFormer} & \multirow{2}{*}{LOcc-BEVDet} & \multirow{2}{*}{LOcc-BEVDet4D} & \multirow{2}{*}{Pseudo-labeled GT} \\
			& & & & Voting voxelization & \multicolumn{1}{c|}{Nearest voxelization} & & & & &  \\
			\hline
			(a) & \checkmark & & & \checkmark & &  & 15.04 & 17.56 & 19.37 & 22.12 \\
			(b) & & \checkmark & & \checkmark & &  & 17.87 & 19.61 & 22.15 & 22.66 \\\hline
			\rowcolor{gray!20} (c) & & & \checkmark & \checkmark &  &  & \textbf{18.62} & \textbf{20.29} & \textbf{22.95} & \textbf{25.53} \\
			\hline
			(d) & & & \checkmark & & \checkmark & & 18.32 & 19.98 & 22.68 & 23.80 \\
			(e) & & & \checkmark & & & \checkmark & 13.77 & 14.49 & 15.41 & 19.55 \\
			\bottomrule
		\end{tabular}
	}
	\label{ablation:pipeline}
\end{table*}

\section{Experiments}
\label{sec:experiments}
\subsection{Datasets and Metrics}
\textbf{Datasets.} 
We evaluate LOcc on the nuScenes dataset~\cite{nuScenes, occ3d}, a large-scale autonomous driving dataset collected across various cities and weather conditions. Each frame includes a LiDAR scan and six RGB images from different viewpoints. It contains 700 training scenes and 150 validation scenes. The spatial volume ranges from $-\text{40}m$ to $\text{40}m$ for the X and Y axes, and $-\text{1}m$ to $\text{5.4}m$ for the Z axis. Voxelization of this volume produces a set of 3D grids with a resolution of $\text{200}\times \text{200}\times \text{16}$, where each voxel measures
$\text{0.4}m\times \text{0.4}m\times \text{0.4}m$. The ground truth includes 17 unique labels (16 semantic classes and 1 free class). 

\noindent\textbf{Metrics.} Following previous approaches~\cite{occ3d}, we list the intersection over union (IoU) and mean IoU (mIoU) metrics for occupied voxel grids and voxel-wise semantic predictions, respectively, providing a comprehensive analysis for  geometry and semantic aspects of the scene.

\subsection{Evaluation Benchmarks}
The evaluation benchmarks focus on two settings, i.e., zero-shot and open-vocabulary. In the zero-shot setting, none of the ground truth classes are seen during training. In contrast, the open-vocabulary setting assumes that only a subset of classes is available during training, and the model is expected to recognize both seen and unseen classes during evaluation

\definecolor{nbarrier}{RGB}{112, 128, 144}
\definecolor{nbicycle}{RGB}{220, 20, 60}
\definecolor{nbus}{RGB}{255, 127, 80}
\definecolor{ncar}{RGB}{255, 158, 0}
\definecolor{nconstruct}{RGB}{233, 150, 70}
\definecolor{nmotor}{RGB}{255, 61, 99}
\definecolor{npedestrian}{RGB}{0, 0, 230}
\definecolor{ntraffic}{RGB}{47, 79, 79}
\definecolor{ntrailer}{RGB}{255, 140, 0}
\definecolor{ntruck}{RGB}{255, 99, 71}
\definecolor{ndriveable}{RGB}{0, 207, 191}
\definecolor{nother}{RGB}{175, 0, 75}
\definecolor{nsidewalk}{RGB}{75, 0, 75}
\definecolor{nterrain}{RGB}{112, 180, 60}
\definecolor{nmanmade}{RGB}{222, 184, 135}
\definecolor{nvegetation}{RGB}{0, 175, 0}
\definecolor{nothers}{RGB}{0, 0, 0}

\begin{table*}
	\footnotesize
	\caption{Detailed metrics of the pseudo-labeled ground truth from different generation settings on the Occ3D-nuScenes dataset. The settings match those outlined in Table~\ref{ablation:pipeline}. We report the mean IoU (mIoU) for semantics across different categories. The best results of different settings are in \textbf{bold}.}
	\begin{center}
		\setlength{\tabcolsep}{0.0057\linewidth}
		\resizebox{\textwidth}{!}{\begin{tabular}{l|c| c c c c c c c c c c c c c c c c c}
				\toprule
				Setting & mIoU
				& \rotatebox{90}{\textcolor{nothers}{$\blacksquare$} others}
				
				& \rotatebox{90}{\textcolor{nbarrier}{$\blacksquare$} barrier}
				
				& \rotatebox{90}{\textcolor{nbicycle}{$\blacksquare$} bicycle}
				
				& \rotatebox{90}{\textcolor{nbus}{$\blacksquare$} bus}
				
				& \rotatebox{90}{\textcolor{ncar}{$\blacksquare$} car}
				
				& \rotatebox{90}{\textcolor{nconstruct}{$\blacksquare$} const. veh.}
				
				& \rotatebox{90}{\textcolor{nmotor}{$\blacksquare$} motorcycle}
				
				& \rotatebox{90}{\textcolor{npedestrian}{$\blacksquare$} pedestrian}
				
				& \rotatebox{90}{\textcolor{ntraffic}{$\blacksquare$} traffic cone}
				
				& \rotatebox{90}{\textcolor{ntrailer}{$\blacksquare$} trailer}
				
				& \rotatebox{90}{\textcolor{ntruck}{$\blacksquare$} truck}
				
				& \rotatebox{90}{\textcolor{ndriveable}{$\blacksquare$} drive. suf.}
				
				& \rotatebox{90}{\textcolor{nother}{$\blacksquare$} other flat}
				
				& \rotatebox{90}{\textcolor{nsidewalk}{$\blacksquare$} sidewalk}
				
				& \rotatebox{90}{\textcolor{nterrain}{$\blacksquare$} terrain}
				
				& \rotatebox{90}{\textcolor{nmanmade}{$\blacksquare$} manmade}
				
				& \rotatebox{90}{\textcolor{nvegetation}{$\blacksquare$} vegetation}
				\\
				\midrule
				(a) & 22.12 & 0.00 & \textbf{10.20} & 7.18 & 29.59 & 35.22 & \textbf{15.70} & 12.80 & 11.19 & 1.86 & \textbf{15.28} & 21.56 & 58.52 & 0.00 & 25.80 & 22.72 & 55.53 & 52.92 \\
				(b) & 22.66 & 0.00 & 8.07 & 4.51 & 23.33 & 37.48 & 9.62 & 11.42 & 14.04 & 10.50 & 4.61 & 19.86 & 60.30 & 0.24 & 30.94 & 25.97 & 59.17 & 65.22 \\\hline
				\rowcolor{gray!20}(c) & \textbf{25.53} & 0.00 & 9.40 & \textbf{8.47} & \textbf{30.16} & \textbf{38.00} & 11.43 & \textbf{20.20} & \textbf{11.39} & \textbf{11.54} & 9.55 & \textbf{25.73} & \textbf{61.86} & 0.40 & \textbf{36.41} & \textbf{33.00} & \textbf{59.42} & \textbf{67.10 }\\\hline
				(d) & 23.80 & 0.00 & 8.90 & 6.75 & 27.86 & 34.21 & 10.19 & 15.75 & 10.05 & 8.68 & 8.83 & 22.67 & 60.91 & \textbf{0.50} & 34.91 & 31.82 & 56.91 & 65.70 \\
				(e) & 19.55 & 0.00 & 7.19 & 4.28 & 24.62 & 21.15 & 8.30 & 8.43 & 6.71 & 4.63 & 8.65 & 16.01 & 55.41 & \textbf{0.50} & 26.32 & 31.01 & 50.95 & 58.22 \\
				\bottomrule
			\end{tabular}
		}
	\end{center}
	\label{ablation:recon_gt}
\end{table*}

a
\subsection{Implementation Details}
\noindent\textbf{Network Structures.} To demonstrate the generalizability of the proposed LOcc, we evaluate it on BEVFormer~\cite{bevformer}, BEVDet~\cite{BEVDet}, and BEVDet4D~\cite{BEVDet4D}, including both forward- and backward-projection methods. For BEVDet~\cite{BEVDet} and BEVDet4D~\cite{BEVDet4D}, we employ ResNet-50~\cite{ResNet} as the image backbone. The resolution of the voxel features after the 2D-to-3D transformation is $\text{200}\times \text{200}\times \text{16}$, with a feature dimension of 64. For BEVFormer~\cite{bevformer}, we adopt ResNet101-DCN~\cite{ResNet} as the backbone, with the default size for BEV queries set to $\text{200}\times \text{200}$, with a feature dimension of 256. We remove the temporal self-attention layer and use 6 cross-attention layers. For each query, it corresponds to 4 target points with different heights in 3D space, with height anchors uniformly predefined from $-\text{1}m$ to $\text{5.4}m$. The number of sampling points around each reference point is set to 8. For all models, the image size is resized to either  $\text{256}\times \text{704}$ or $\text{512}\times \text{1408}$, and the language features are compressed to a dimension of 128.

\noindent\textbf{Training Setup.}  We implement the networks using PyTorch. The models are trained for 25 epochs on 8 NVIDIA 3090 GPUs, with a batch size of 1 on each GPU. We employ the AdamW~\cite{AdamW} optimizer with $\beta_{1}=\text{0.9}, \beta_{2}=\text{0.99}$. The maximum learning rate is set to $\text{3}\times\text{10}^{-\text{4}}$, and the cosine annealing learning rate strategy is adopted for the learning rate decay, where the cosine warm up strategy is applied for the first $\text{5}\%$ iterations.

\begin{figure*}
	\centering
	\newcolumntype{P}[1]{>{\centering\arraybackslash}m{#1}}
	\renewcommand{\arraystretch}{1.0}
	\footnotesize
	\begin{tabular}{P{0.17\textwidth} P{0.17\textwidth} P{0.17\textwidth} P{0.17\textwidth} P{0.17\textwidth}}	
		\includegraphics[width=.9\linewidth]{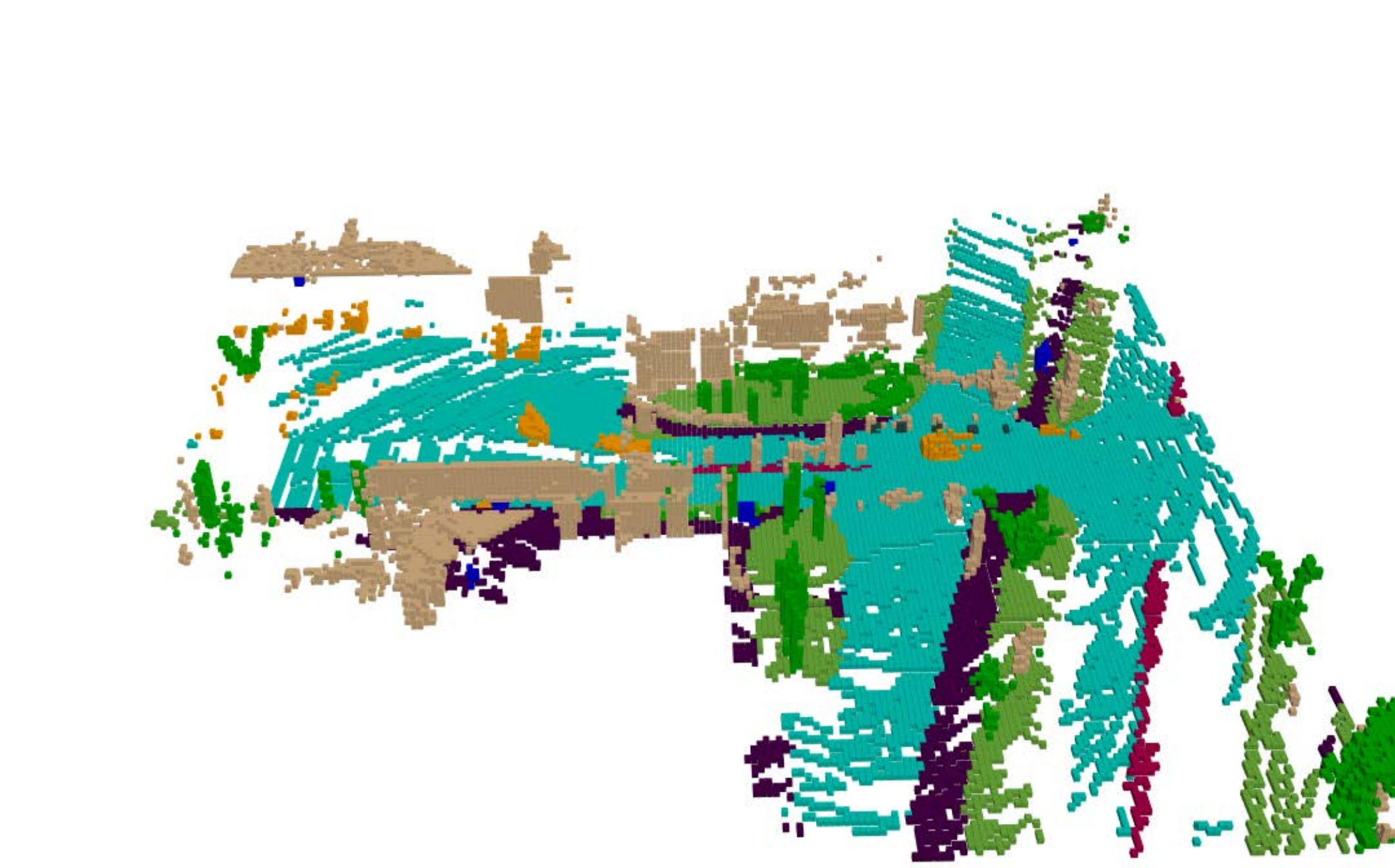} &
		\includegraphics[width=.9\linewidth]{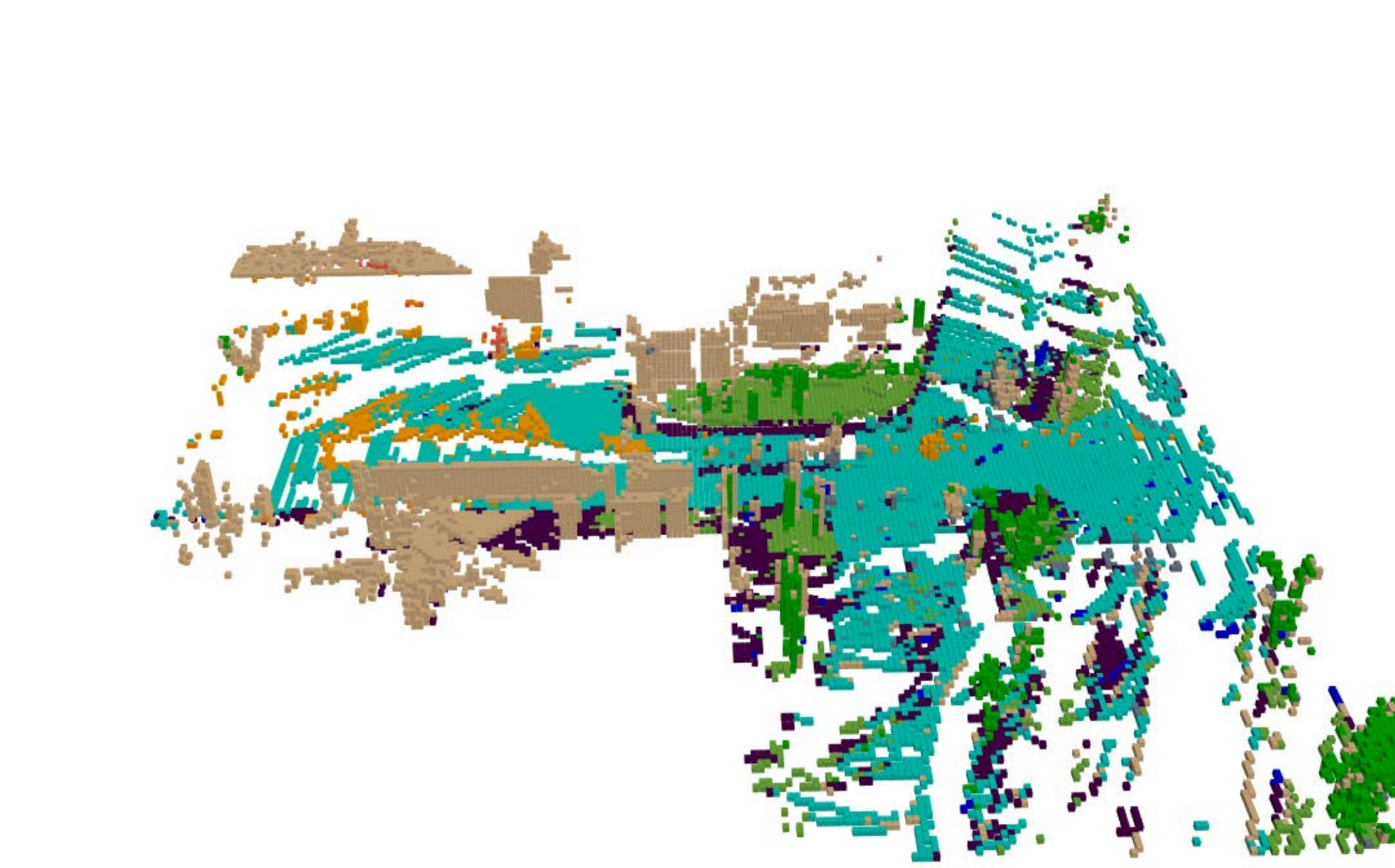} &
		\includegraphics[width=.9\linewidth]{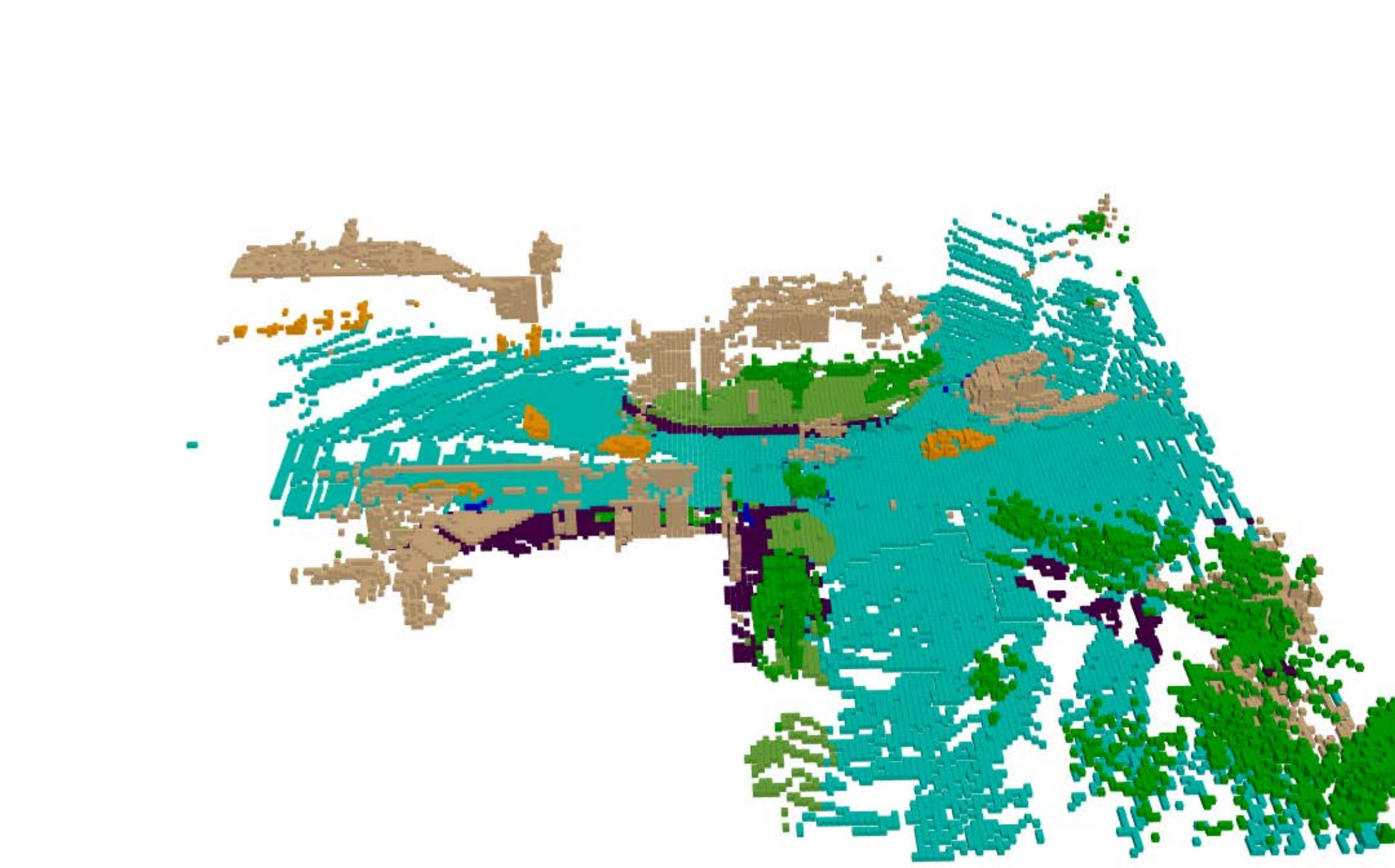} & 
		\includegraphics[width=.9\linewidth]{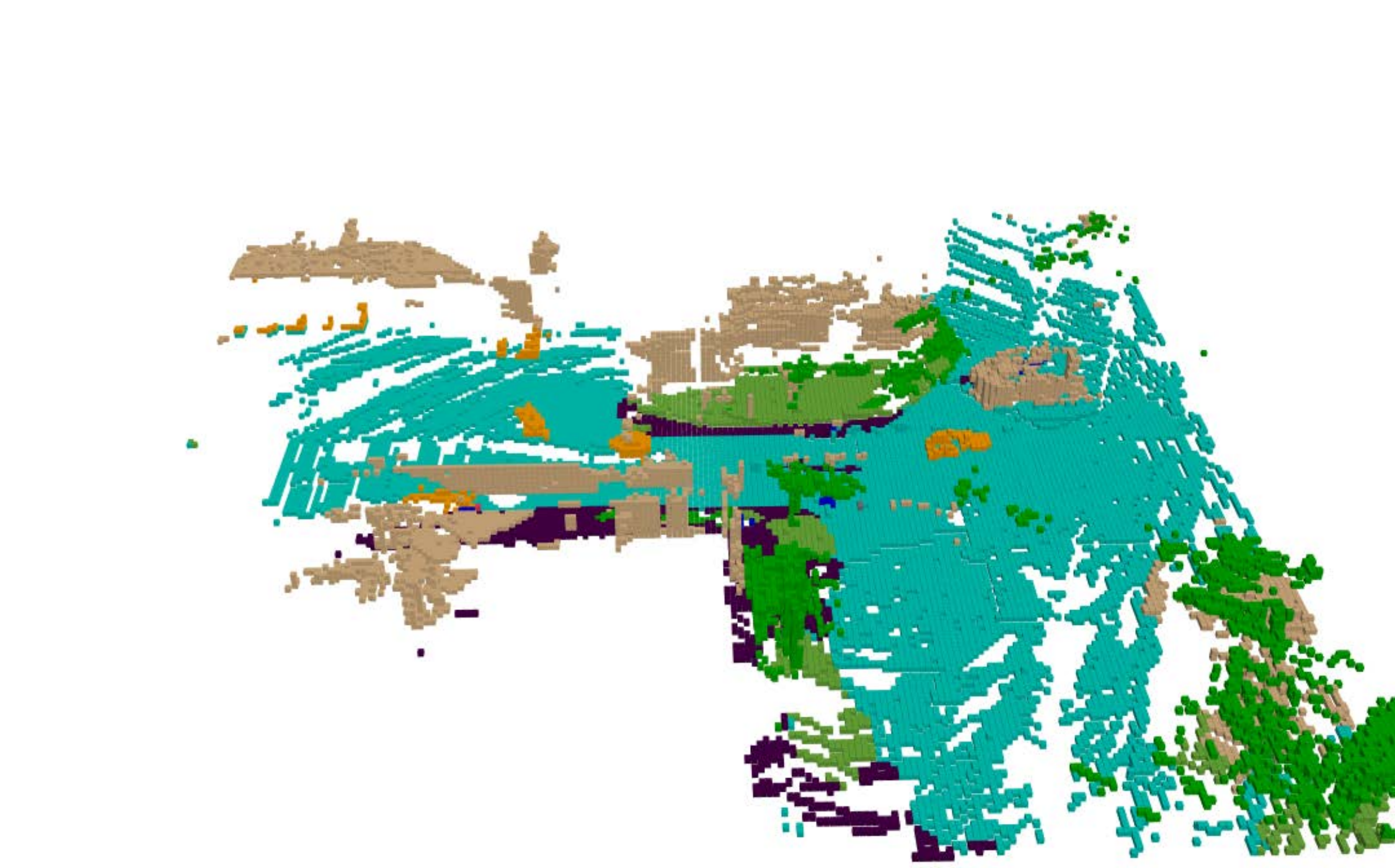} & 
		\includegraphics[width=.9\linewidth]{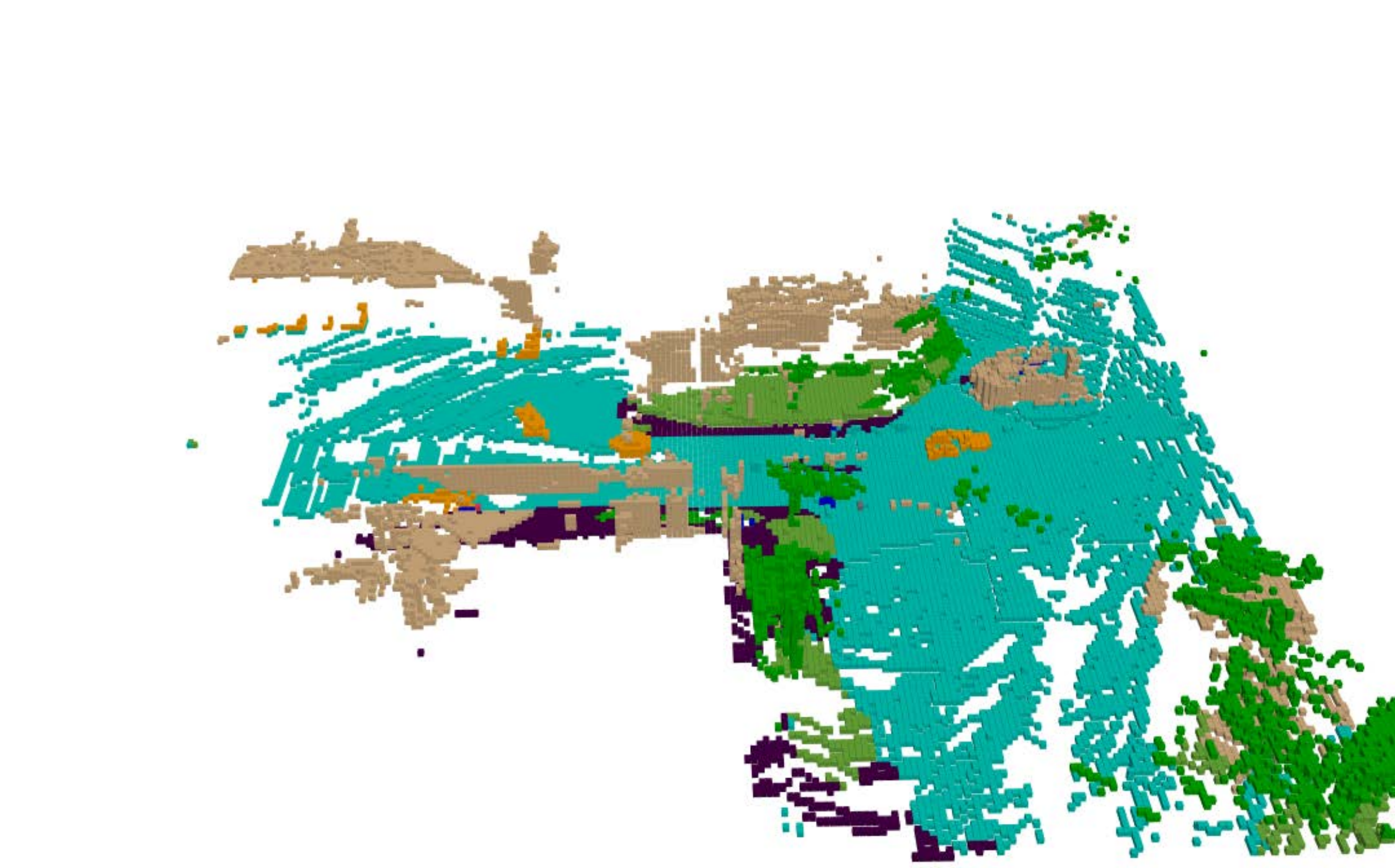}
		\\[-0.1em]
		\includegraphics[width=.9\linewidth]{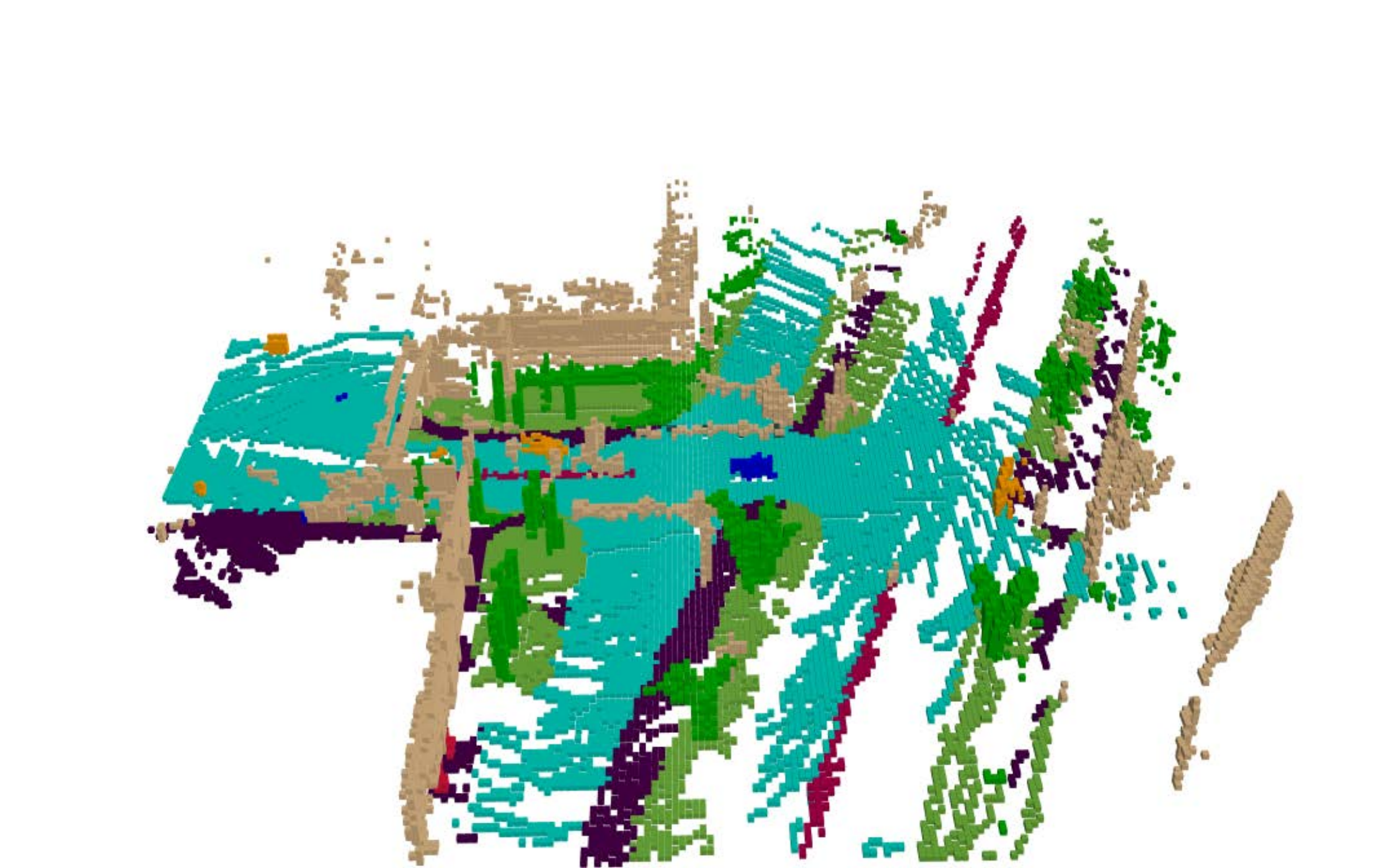} &
		\includegraphics[width=.9\linewidth]{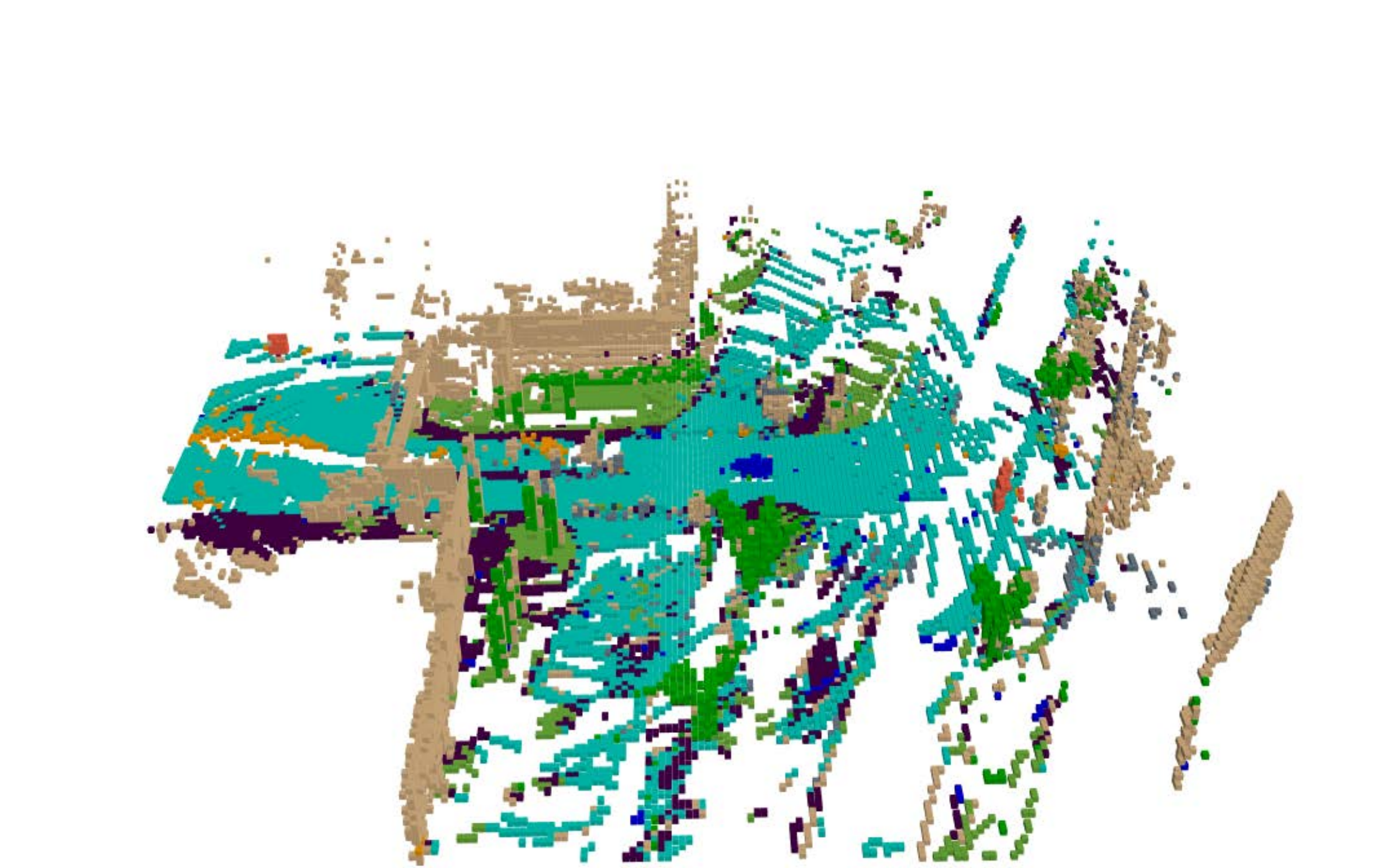} &
		\includegraphics[width=.9\linewidth]{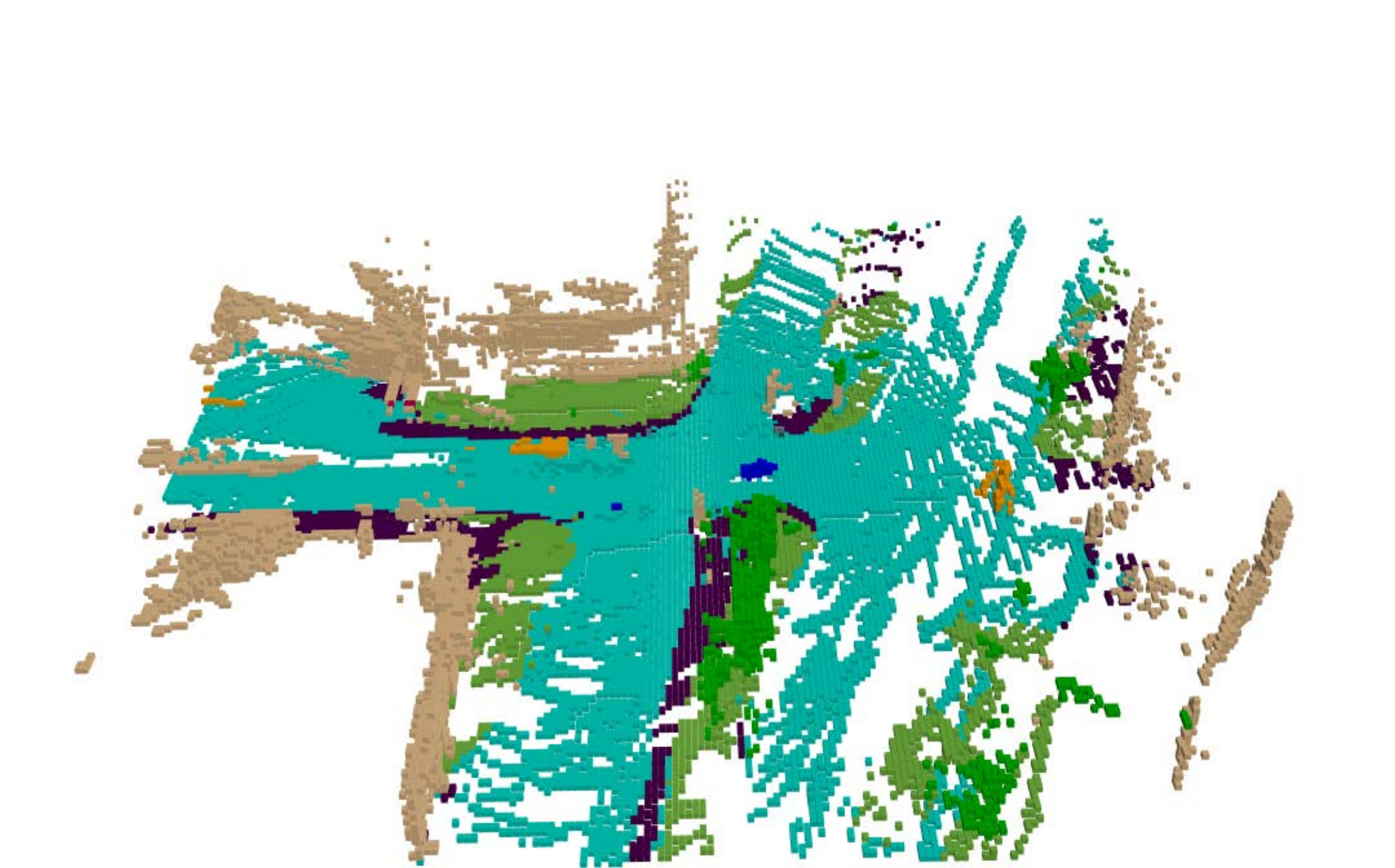} & 
		\includegraphics[width=.9\linewidth]{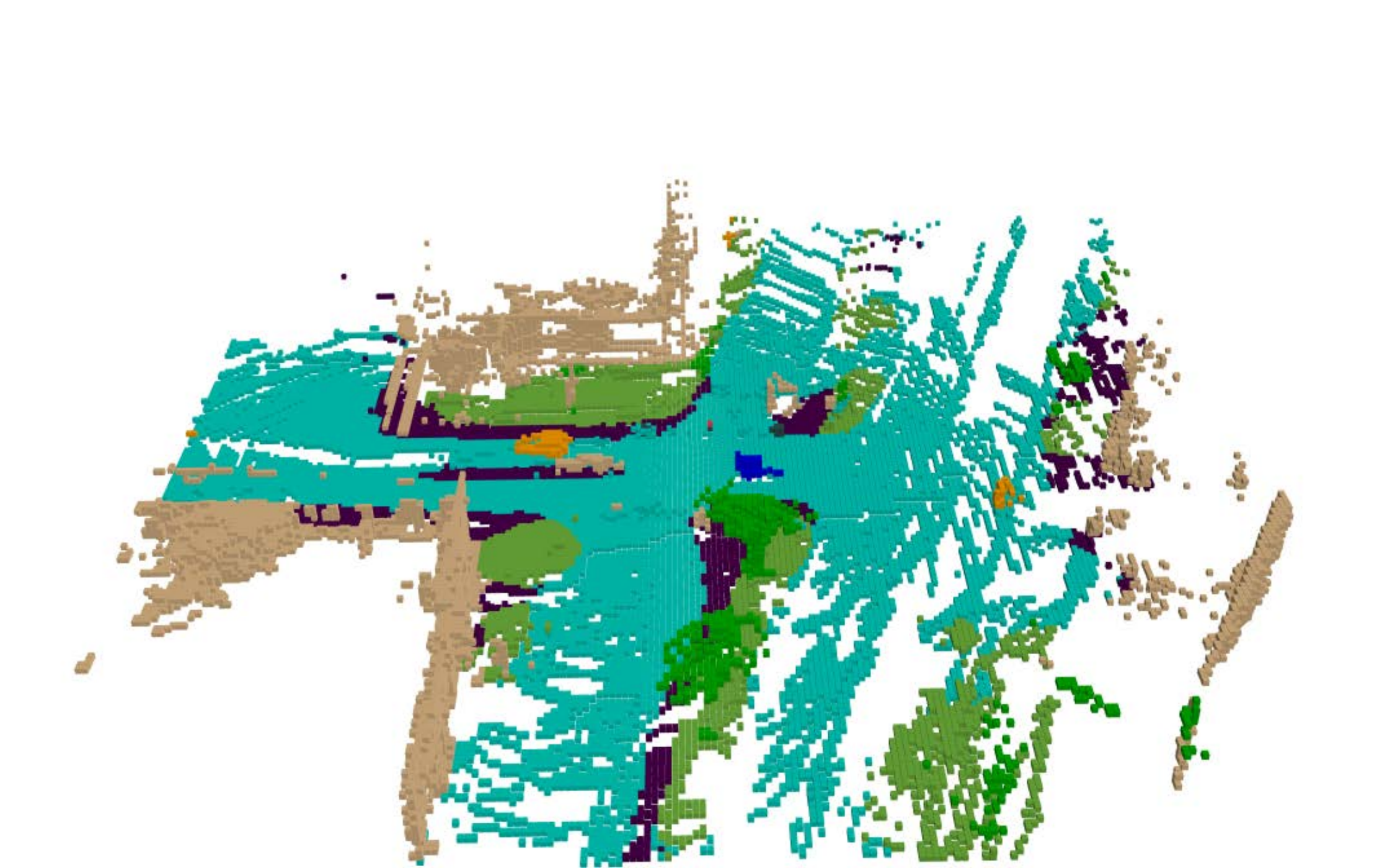} &
		\includegraphics[width=.9\linewidth]{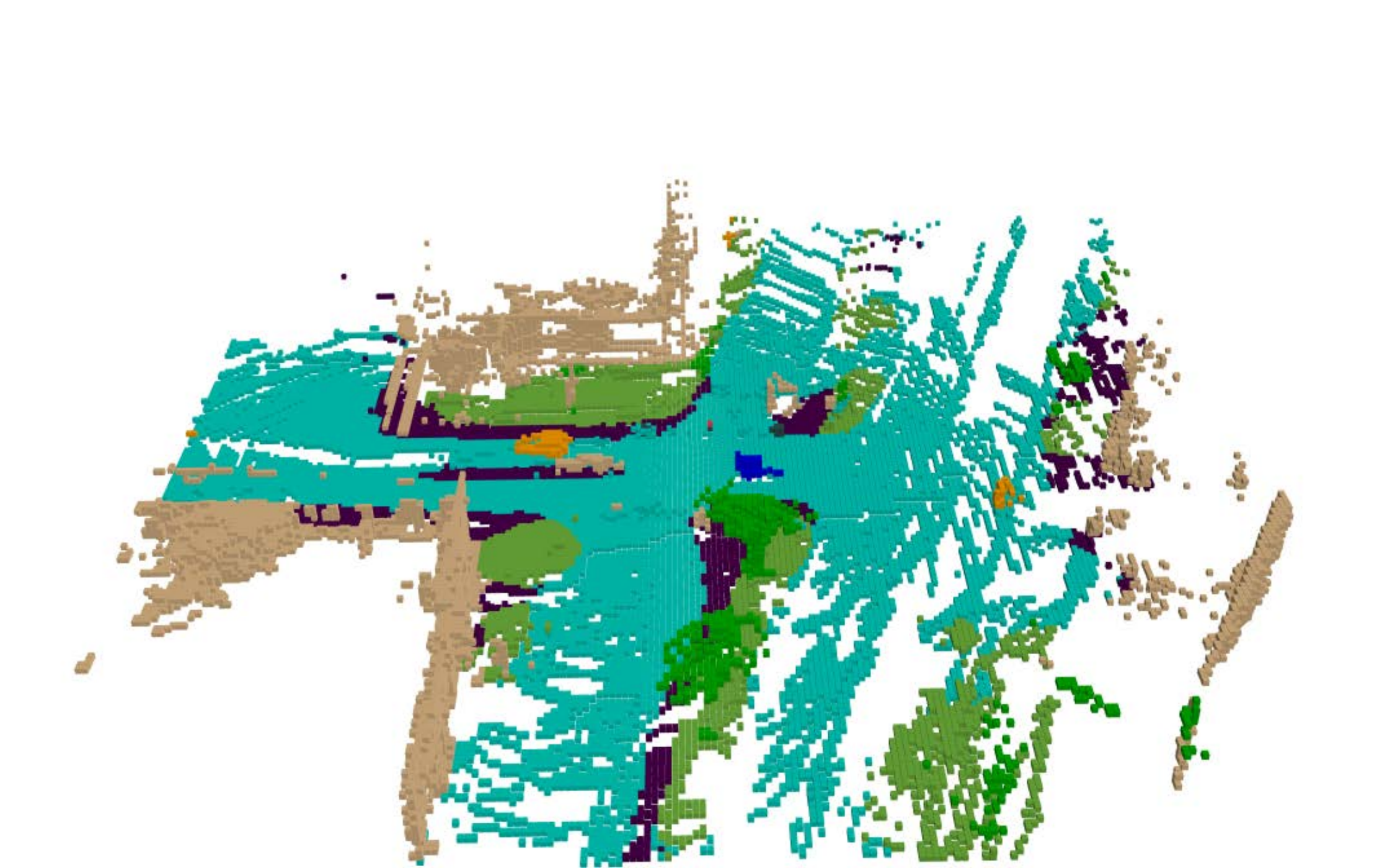}
		\\[-0.1em]
	(a) Ground Truth & (b) Pseudo Ground Truth & (c) LOcc-BEVFormer & (d) LOcc-BEVDet & (e) LOcc-BEVDet4D
	\end{tabular}
	\caption{Quantitative visualization results on the Occ3D-nuScenes~\cite{occ3d} dataset.}
	\label{fig:qualitative_results}
\end{figure*} 
\definecolor{nbarrier}{RGB}{112, 128, 144}
\definecolor{nbicycle}{RGB}{220, 20, 60}
\definecolor{nbus}{RGB}{255, 127, 80}
\definecolor{ncar}{RGB}{255, 158, 0}
\definecolor{nconstruct}{RGB}{233, 150, 70}
\definecolor{nmotor}{RGB}{255, 61, 99}
\definecolor{npedestrian}{RGB}{0, 0, 230}
\definecolor{ntraffic}{RGB}{47, 79, 79}
\definecolor{ntrailer}{RGB}{255, 140, 0}
\definecolor{ntruck}{RGB}{255, 99, 71}
\definecolor{ndriveable}{RGB}{0, 207, 191}
\definecolor{nother}{RGB}{175, 0, 75}
\definecolor{nsidewalk}{RGB}{75, 0, 75}
\definecolor{nterrain}{RGB}{112, 180, 60}
\definecolor{nmanmade}{RGB}{222, 184, 135}
\definecolor{nvegetation}{RGB}{0, 175, 0}
\definecolor{nothers}{RGB}{0, 0, 0}

\begin{table*}
	\footnotesize
	\caption{Open-vocabulary occupancy prediction results on the Occ3D-nuScenes dataset. We report the mean IoU (mIoU) for semantics across different categories, along with per-class semantic IoUs. The best results among the methods are in \textbf{bold}.}
	\begin{center}
		\setlength{\tabcolsep}{0.0057\linewidth}
		\resizebox{\textwidth}{!}{\begin{tabular}{l|c|c|c|cccc|ccccccccccccc}
				\toprule
				Model 
				& Image Backbone & Image Size
				& mIoU 
				& \rotatebox{90}{\textcolor{nbicycle}{$\blacksquare$} bicycle}
				
				& \rotatebox{90}{\textcolor{nmotor}{$\blacksquare$} motorcycle}
				
				& \rotatebox{90}{\textcolor{ntraffic}{$\blacksquare$} traffic cone}
				
				& \rotatebox{90}{\textcolor{nsidewalk}{$\blacksquare$} sidewalk}
				
				& \rotatebox{90}{\textcolor{nothers}{$\blacksquare$} others}
				
				& \rotatebox{90}{\textcolor{nbarrier}{$\blacksquare$} barrier}
				
				& \rotatebox{90}{\textcolor{nbus}{$\blacksquare$} bus}
				
				& \rotatebox{90}{\textcolor{ncar}{$\blacksquare$} car}
				
				& \rotatebox{90}{\textcolor{nconstruct}{$\blacksquare$} const. veh.}
				
				& \rotatebox{90}{\textcolor{npedestrian}{$\blacksquare$} pedestrian}
				
				& \rotatebox{90}{\textcolor{ntrailer}{$\blacksquare$} trailer}
				
				& \rotatebox{90}{\textcolor{ntruck}{$\blacksquare$} truck}
				
				& \rotatebox{90}{\textcolor{ndriveable}{$\blacksquare$} drive. suf.}
				
				& \rotatebox{90}{\textcolor{nother}{$\blacksquare$} other flat}

				& \rotatebox{90}{\textcolor{nterrain}{$\blacksquare$} terrain}
				
				& \rotatebox{90}{\textcolor{nmanmade}{$\blacksquare$} manmade}
				
				& \rotatebox{90}{\textcolor{nvegetation}{$\blacksquare$} vegetation}
				\\
				\midrule
				LOcc-BEVFormer (ours) & ResNet-101 & 256 $\times$ 704 & 23.26 & 18.71 & 22.21 & 24.33 & 48.14 & 0.00 & 14.48 & 28.16 & 33.88 & 4.32 & 9.19 & 2.37 & 22.45 & 72.09 & 0.00 & 34.32 & 31.22 & 29.58 \\
				LOcc-BEVDet (ours) & ResNet-50 & 256 $\times$ 704 & 23.50 & 12.21 & 17.33 & 17.24 & 47.03 & 0.00 & 13.43 & 32.67 & 35.54 & 8.68 & 10.01 & 2.54 & 27.33 & 72.16 & 0.00 & 36.31 & 34.47 & 32.57 \\
				LOcc-BEVDet4D (ours) & ResNet-50 & 256 $\times$ 704 & 26.91 & 20.34 & 23.77 & 23.93 & 51.47 & 0.00 & 15.63 & 29.54 & 39.89 & 10.99 & 14.01 & 6.21 & 31.72 & 73.23 & 0.00 & 40.11 & 39.00 & 37.76 \\
				\bottomrule
			\end{tabular}
		}
	\end{center}
	\label{tab:open_vocabulary}
	\vspace{-3mm}
\end{table*}
\begin{table}[t]
	\centering
	\Large
	\caption{Ablation study on the OV-Seg models. The best results among different methods are in \textbf{bold}.}
	\resizebox{\linewidth}{!}{
		\begin{tabular}{c|ccc}
			\toprule
			\multirow{2}{*}{Method} & \multicolumn{3}{c}{mIoU} \\\cline{2-4}
			& LOcc-BEVDet & LOcc-BEVDet4D & LOcc-BEVFormer \\\hline
			ODISE~\cite{ODISE} & 20.17 & 22.52 & \textbf{18.69} \\
			CAT-Seg~\cite{Catseg} & \textbf{20.68} & \textbf{22.76} & 18.57 \\
			SAN~\cite{SAN} & 20.29 & 22.68 & 18.62 \\
			\bottomrule
		\end{tabular}
	}
	\label{ablation:open_seg}

\end{table}

\subsection{Zero-shot Occupancy Prediction}
We evaluate the effectiveness of the proposed LOcc framework on three models: BEVFormer~\cite{bevformer}, BEVDet~\cite{BEVDet} and BEVDet4D~\cite{BEVDet4D}, referred to as LOcc-BEVFormer, LOcc-BEVDet, and LOcc-BEVDet4D, respectively. The comparison methods includes those that rely solely on images~\cite{SelfOcc, OccNeRF} and those that utilize unlabeled LiDAR data to assist in geometry prediction~\cite{pop3d, VEON}. To ensure a fair comparison with POP-3D, which uses single-frame LiDAR for geometry supervision, we replace the single-frame LiDAR data with dense binary occupancy ground truth~\cite{VEON}. For OV-Seg models, we replace LSeg~\cite{LSeg} with more powerful SAN~\cite{SAN}. 

Table~\ref{tab:zero_shot} displays detailed results of zero-shot occupancy prediction. With the input image of size $\text{256}\times\text{704}$, LOcc-BEVFormer, LOcc-BEVDet, and LOcc-BEVDet4D achieve an mIoU of 18.62, 20.29, and 22.95, respectively. Notably, LOcc-BEVDet, a more simpler model, surpasses all previous approaches, even though many of these approaches use larger image backbones or higher-resolution inputs. Furthermore, when the image resolution is increased to $\text{512}\times\text{1408}$, the models attain an better mIoU of 19.85, 20.84, and 23.84.

\subsection{Ablation Study}
We conduct ablation study to evaluate the effectiveness of the proposed LOcc framework. 

\textbf{Label Transferring Process.} 
Table~\ref{ablation:pipeline} provides an analysis of the label transferring process. First, we replace the segmentation maps with image features extracted from SAN~\cite{SAN}, requiring the learned 3D features to align with these image features. This configuration indirectly establishes correspondences with textual labels. Setting (a) demonstrates that it leads to a performance drop in the generated pseudo-labeled ground truth, which in turn degrades the performance of all models. Next, we apply our proposed semantic transitive labeling pipeline using only single-frame vocabularies. As shown in setting (b), the mIoU of the pseudo-labeled ground truth increases to 22.66, and the metrics for all models improve to 17.87, 19.61, and 22.15, respectively. Finally, by treating multiple consecutive frames as a unified sequence and consolidating the vocabulary from each frame into a cohesive set for pseudo-label assignment, as indicated in setting (c), we observe additional performance gains, further validating the effectiveness of our label transferring process.

\textbf{Scene Reconstruction Process.} The voxel-based model-view projection treats all voxels as point clouds, making it challenging to determine whether a voxel lies on the surface after projection. This coarse approach overlooks the occlusion problem, leading to noisy voxel-to-text correspondences. Setting (e) demonstrates that the pseudo-labeled ground truth from voxel-based model-view projection can only achieve an mIoU of 19.55. Using this unsatisfied ground truth to train the OVO models results in a huge performance drop to 13.77, 14.49, and 15.41. Instead, each frame of LiDAR can be accurately calibrated with corresponding images during the data collection process, significantly reducing misprojections. Thus, performing scene reconstruction with nearest-point voxelization yields notable improvements, as shown in setting (d). Furthermore, majority-voting voxelization can mitigate the issue of wrong segmentation results, achieving better results with the mIoU of pseudo-labeled and models improving to 25.53, 18.62, 20.29, and 22.68, as illustrated in setting (c). These results prove that majority-voting is more robust and less susceptible to the influence of single-point projection errors. The overall ablation results prove that the quality of the pseudo-labeled ground truth highly influences the performance of OVO models. In Table~\ref{ablation:recon_gt}, we present detailed metrics for the pseudo-labeled ground truth generated under various settings. Consistent with prior analyses, our pipeline yields better results on most of categories, further confirming the effectiveness of our proposed pipeline.

\textbf{OV-Seg Models.} Table~\ref{ablation:open_seg} lists the results of the OVO networks trained on the pseudo-labeled ground truth generated from different OV-Seg models: SAN~\cite{SAN}, ODISE~\cite{ODISE} and CAT-Seg~\cite{Catseg}.  In this work, we use SAN~\cite{SAN} for fair comparison with previous approaches~\cite{VEON}. 

\subsection{Open-vocabulary Occupancy Prediction}
Table~\ref{tab:open_vocabulary} lists the detailed metrics of open vocabulary occupancy prediction. We use \textit{bicycle}, \textit{motorcycle}, \textit{traffic cone}, and \textit{sidewalk} as base classes, while treating the remaining categories as novel classes. With more seen classes, the overall performance for the models rise. Besides, the IoUs on unseen classes are always competitive, proving the open-vocabulary capability of our methods.
\subsection{Qualitative Results}
Fig.~\ref{fig:qualitative_results} presents visualizations of the pseudo-labeled ground truth, along with results from LOcc-BEVFormer, LOcc-BEVDet, and LOcc-BEVDet4D trained on it. Each model demonstrates clear predictions, despite the noise present in the training dataset. This demonstrates the ability of the models to learn useful information from semi-labeled data. These findings are also observed in other vision tasks, such as image segmentation~\cite{SAM}, video segmentation~\cite{SAM2} ,and depth estimation~\cite{DepthAnything}, 
	\section{Conclusions}
\label{sec:conlusions}
In this paper, we have presented LOcc, an effective and generalizable framework for open-vocabulary occupancy prediction. We propose a semantic transitive labeling pipeline for generating dense and fine-grained 3D language occupancy ground truth. We experimentally demonstrates that text semantic labels can be effectively transferred from images to LiDAR point clouds and ultimately to voxels, diminishing labor-intensive human annotations. By replacing the original prediction head with a geometry head and a language head, LOcc is compatible with most existing supervised networks. Based on the BEVFormer, BEVDet, and BEVDet4D, our propose method can achieve better results than all the previous state-of-the-art zero-shot occupancy prediction methods.
	\newpage
\section*{Acknowledgments}
We thank the reviewers for the valuable discussions. This research was supported by the National Key Research and Development Program of China under grant 2023YFB3209800, in part by the National Natural Science Foundation of China under grant 62301484, in part by the Ningbo Natural Science Foundation of China under grant 2024J454, in part by the Aeronautical Science Foundation of China under grant 2024M071076001 and in part by the Zhejiang Provincial Natural Science Foundation of China under Grant No. LD24F030001.
{
    \small
    \bibliographystyle{ieeenat_fullname}
    \bibliography{main}
}

\end{document}